\documentclass{article}

\usepackage[nonatbib,preprint]{neurips_2021}

\usepackage[square,sort,comma,numbers]{natbib}

\usepackage[utf8]{inputenc} % allow utf-8 input
\usepackage[T1]{fontenc}    % use 8-bit T1 fonts
\usepackage{hyperref}       % hyperlinks
\hypersetup{
    colorlinks=true,
    linkcolor=blue,
    filecolor=magenta,      
    urlcolor=blue,
}
\usepackage{url}            % simple URL typesetting
\usepackage{booktabs}       % professional-quality tables
\usepackage{amsfonts}       % blackboard math symbols
\usepackage{nicefrac}       % compact symbols for 1/2, etc.
\usepackage{microtype}      % microtypography
\usepackage[table,xcdraw]{xcolor} % Tables and Color
\usepackage{graphicx}
\usepackage{caption}
\usepackage{subcaption}
\usepackage{footnote}
\usepackage{booktabs}
\usepackage{algorithm}
\usepackage{algorithmic}
\usepackage{wrapfig}
\usepackage{enumitem}
\usepackage{amssymb,amsmath}
\usepackage{mathtools}
\usepackage{multirow}
\usepackage[normalem]{ulem} % Underline
\usepackage{comment}
\usepackage{longtable}
\usepackage{todonotes}

\graphicspath{ {./images/} }
\title{Modern Evolution Strategies for Creativity:\\ Fitting Concrete Images and Abstract Concepts}

\author{
  Yingtao Tian \\
  Google Brain\\
  \texttt{alantian@google.com} \\
  \And
  David Ha\\
  Google Brain\\
  \texttt{hadavid@google.com} \\
}

\date{} % Omit date

\begin{document}

\maketitle

\vskip -0.4cm % useful knobs to optimize layout. Use cm for consistency throughout the manuscript.
\begin{abstract}
\vskip -0.3cm
Evolutionary algorithms have been used in the digital art scene since the 1970s. A popular application of genetic algorithms is to optimize the procedural placement of vector graphic primitives to resemble a given painting. In recent years, deep learning-based approaches have also been proposed to generate procedural drawings, which can be optimized using gradient descent. In this work, we revisit the use of evolutionary algorithms for computational creativity. We find that modern evolution strategies (ES) algorithms,  when tasked with the placement of shapes, offer large improvements in both quality and efficiency compared to traditional genetic algorithms, and even comparable to gradient-based methods. We demonstrate that ES is also well suited at optimizing the placement of shapes to fit the CLIP model, and can produce diverse, distinct geometric abstractions that are aligned with human interpretation of language. 
Videos and demo: \url{https://es-clip.github.io/}.
% Videos and demo: https://es-clip.github.io/ % line in the abstract submission form of arxiv.org
\end{abstract}

\begin{figure}[!b] % [!b] instad of [!htb] Force this figure to be at the bottom of the first page.
\centering
\vskip -0.9cm
\includegraphics[trim=0 55px 70px 0, clip, width=0.95\textwidth]{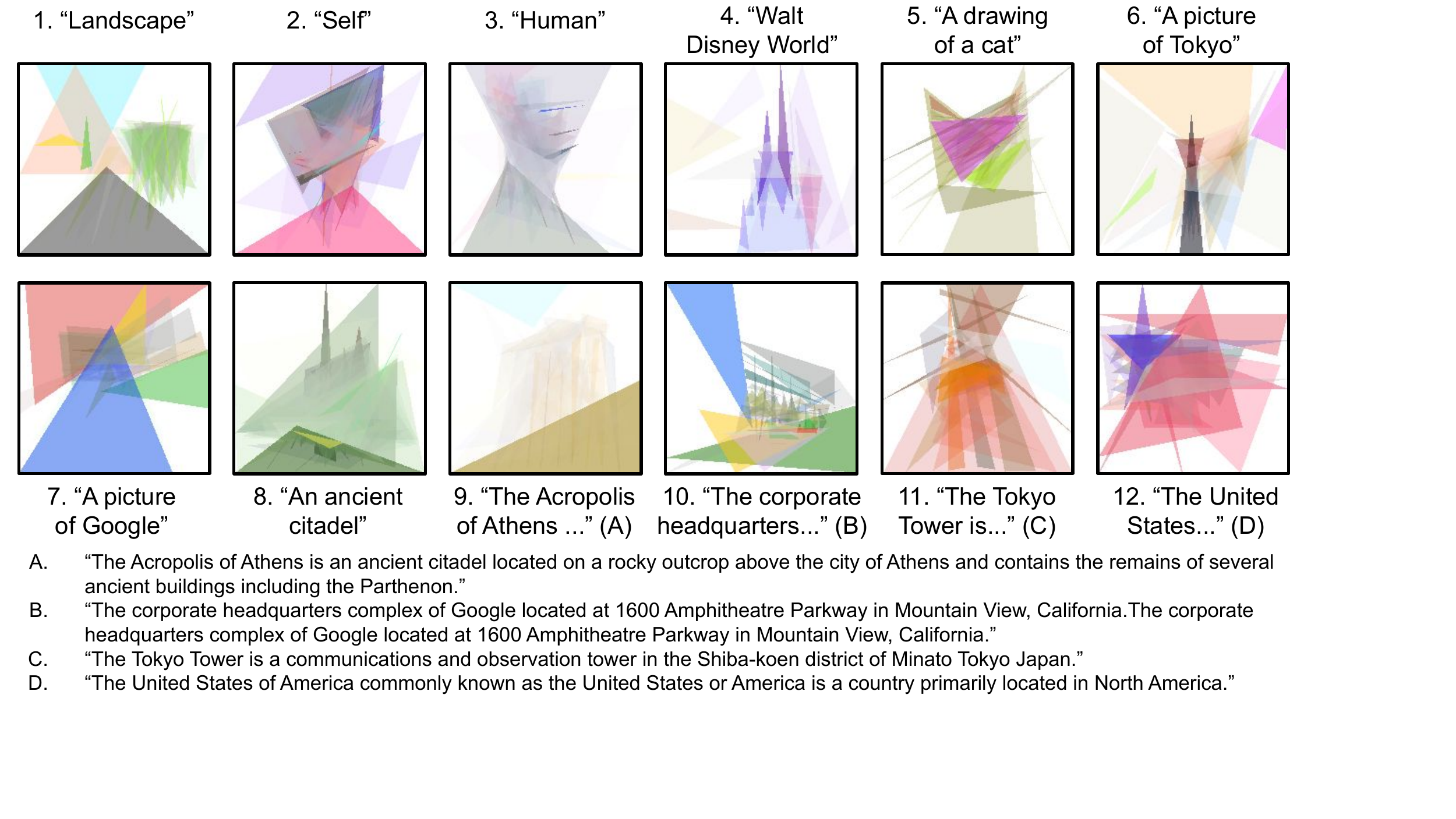}
\vskip -0.1cm
\caption{Our proposed painting synthesization places transparent triangles using evolution strategy (ES). 
Each concept represented as a text prompt is accompanied by its corresponding synthesized image.
Here, the fitness is defined as the cosine distance between rendered canvas and text, both embedded by CLIP~\cite{radford2021learning}, and we optimize the position and the color of triangles using ES.
}
\label{fig:summary}
\vskip -0.4cm
\end{figure}
 % Summary figure should be on the first page.

\begin{figure}[!htb]
\centering
\vskip -0.3cm
\includegraphics[trim=0 180px 200px 0, clip, width=\textwidth]{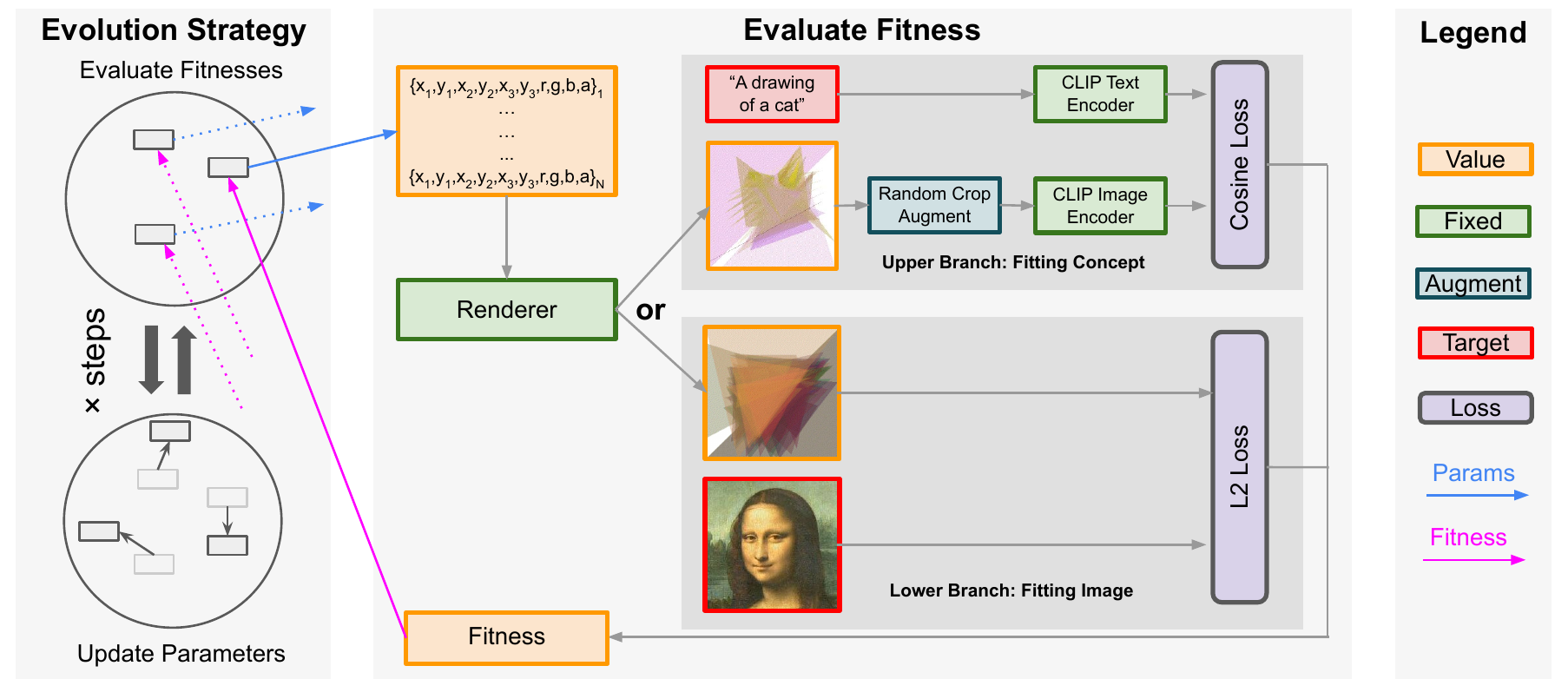}
\vskip -0.1cm
\caption{The architecture of our method. Our proposed method synthesizes painting by placing transparent triangles using Evolution Strategy (ES). 
After rendering the parameters on a canvas, we calculate the fitness, which measures how well the canvas fits either a target image, or a concept in the form of a text prompt. 
The fitness, in turn, guides the evolution process to find better parameters.
}
\vskip -0.5cm
\label{fig:architecture}
\end{figure}

\section{Introduction}
\label{sec:introduction}
\vskip -0.1cm

Staring from early 20th-century in the wider context of modernism~\cite{kuiper2021modernism}, a series of avant-garde art abandoned the depiction of objects from tradition rules of perspective and instead picking revolutionary, abstract point of views.
The Cubism art movement~\cite{rewald2014heilbrunn}, popularized by influential artists including Pablo Picasso, proposed that objects are analyzed by the artist, broken up, and reassembled in an abstract form consisting of geometric representations. 
This naturally develops into the geometric abstraction~\cite{dabrowski2004geometric}, where pioneer abstractionists like Wassily Kandinsky and Piet Mondrian represented the world using composed primitives that are either purely geometric or elementary.
The impact is far-reaching: The use of simple geometry can be seen as one of styles found in abstract expressionism~\cite{paul2004abstract} where artists expressed their subconscious or impulsive feelings.
It also helped shape the minimalist art~\cite{tate_minimalism} and minimalist architecture~\cite{rose1965abc} movements, in which everything is stripped down to its essential quality to achieve simplicity~\cite{bertoni2002minimalist}. 

The idea of minimalist art has also been explored in computer art with a root in mathematical art~\cite{malkevitch2003mathematics}.
Schmidhuber~\cite{schmidhuber1997low} proposed an art form in the 1990s, called low-complexity art, as the minimal art in the computer age that attempts to depict the essence of an object by making use of ideas from algorithmic complexity~\cite{kolmogorov1965three}.
Similarly, algorithmic art~\cite{verostko1994algorithmic} proposed to generate arts using the algorithm designed by the artist.
In a broad sense, algorithmic art could be said to include genetic algorithm where the artist determines the rules governing how images evolves iteratively, which are a popular method applied to approximate images using simple shapes, often producing abstract art style.
As one example, a basic genetic algorithm using evolution has been proposed~\cite{johansson2008genetic,alteredqualia2008evolutiongenetic} to represent a target image using semi-transparent, overlapping triangles. %(See ``Basic'' in Figure~\ref{fig:es-ours-vs-basic}).
This approach has gained popularity over the years with the creative coding community, resulting in many sophisticated extensions~\cite{fogleman2016,cason2016,berg2019evolved,paauw2019paintings,shahrabi2020}.
These methods are iterative, enabling the creation process~\cite{tate2021process} to be captured.

\begin{figure}[!htb]
\captionsetup[subfigure]{labelformat=empty}
\centering
\vskip -0.05cm
\begin{subfigure}[t]{0.20\textwidth}
    \centering
    \includegraphics[width=\textwidth]{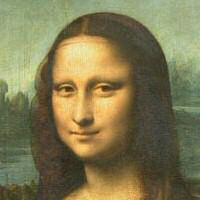}
\end{subfigure}
\hfill
\begin{subfigure}[t]{0.20\textwidth}
    \centering
    \includegraphics[width=\textwidth]{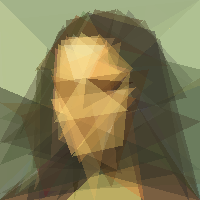}
\end{subfigure}
\hfill
\begin{subfigure}[t]{0.54\textwidth}
    \centering
    \includegraphics[width=\textwidth]{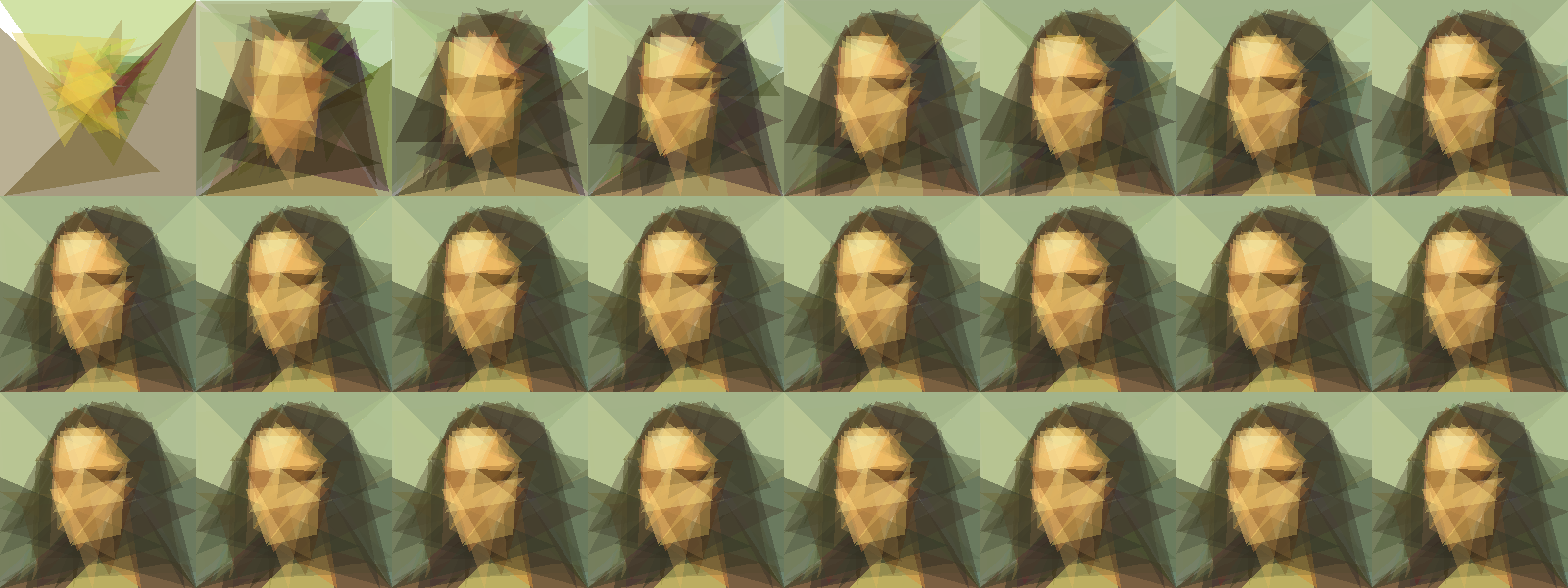}
\end{subfigure}
\vskip -0.2cm
\caption{Our method leverages modern ES (PGPE with ClipUp), with $50$ triangles and runs for $10,000$ steps to fit the target image ``Mona Lisa'' (left). Here it is followed by the  evolved results (middle) and the evolution process (right).}
\label{fig:es-ours-mona-lisa}
\vskip -0.2cm
\end{figure}

With the recent resurgence of interest in evolution strategies (ES) in the machine learning community~\cite{salimans2017evolution,ha2017evolving}, in this work, we revisit the use of ES for creativity applications as an alternative to gradient-based methods. 
For approximating an image with shapes, we find that modern ES algorithms offer large improvements in both quality and efficiency when compared to traditional genetic algorithms, and as we will also demonstrate, even comparable to state-of-the-art differentiable rendering methods~\cite{laine2020modular}.
We show that ES is also well suited at optimizing the placement of shapes to fit the CLIP~\cite{radford2021learning} model, and can produce diverse, distinct geometric abstractions that are aligned with human interpretation of language.
Such an alignment is due to the use of CLIP model that are trained on aligned real-world text-image dataset.
Interestingly, the results produced by our method resemble abstract expressionism~\cite{paul2004abstract} and minimalist art~\cite{tate_minimalism,rose1965abc}.
We provide a reference code implementation of our approach \href{https://es-clip.github.io/}{online} so that it can be a useful tool in the computational artist's toolbox.

\section{Background}
\label{sec:realted-work}
\vskip -0.3cm

\textbf{Related Work}\;In recent years, deep learning has also been applied to methods that can generate procedural drawings, which can be optimized with gradient descent.
A growing list of works~\cite{zheng2018strokenet,nakano2019neural,huang2019learning,liu2021paint} also tackle the problem of approximating pixel images with simulated paint medium, and differentiable rendering~\cite{kato2020differentiable,laine2020modular} methods enable computer graphics to be optimized directly using gradient descent.
To learn abstract representations, probabilistic generative models~\cite{gregor2015draw,ha2017neural,ganin2018synthesizing,mellor2019unsupervised,lopes2019learned} are used to sample procedurally drawings directly from a latent space, without any given input images, similar to their pixel image counterparts.
To interface with natural language, methods have been proposed to procedurally generate drawings of image categories~\cite{white2019shared}, and word embeddings~\cite{huang2019sketchforme,huang2020scones}, enabling an algorithm to \textit{draw what's written}.
This combination of NLP and pixel image generation is explored at larger scale in CLIP~\cite{radford2021learning}, and its procedural sketch counterpart CLIPDraw~\cite{frans2021clipdraw}.

Perhaps among the related works, the closest to our approach is \cite{fernando2021generative}, which, similar to our work, uses a CLIP-like dual-encoder model pre-trained on the ALIGN~\cite{jia2021scaling} dataset to judge the similarity between generated art and text prompt, and leverages evolutionary algorithms to optimize a non-differentiable rendering process. 
However, there are several key differences:
\cite{fernando2021generative} parameterizes the rendering process with a hierarchical neural Lindenmayer system~\cite{lindenmayer1968mathematical} powered by multiple-layer LSTM~\cite{hochreiter1997long} and, as a result, it models well patterns with complex spatial relation, 
whereas our work favors a drastically simpler parameterization which just puts triangles individually on canvas to facilitate a different, minimalist art style that is complementary to theirs~\cite{fernando2021royal}.
Moreover, while \cite{fernando2021generative} uses a simple binary-tournament genetic algorithm~\cite{harvey2009microbial}, we opt for a modern state-of-the-art evolution strategy, PGPE~\cite{sehnke2010parameter} with ClipUp~\cite{toklu2020clipup}, performing well enough to produce interesting results within a few thousand steps.

%\section{Method}
%\label{sec:method}

%In this section, we start with a background part covering evolution strategies (ES) and language-derived image generation, followed by our proposed evolution strategies pipeline for creativity.

\textbf{Evolution Strategies (ES)}~\cite{beyer2001theory,beyer2002evolution} has been applied to optimization problems for a long period of time.
A straightforward implementation of ES can be iteratively perturbing parameters in a pool and keeping those that are most fitting, which is simple yet inefficient. 
As a consequence, applying such a straightforward algorithm can lead to sub-optimal performance for art creativity~\cite{alteredqualia2008evolutiongenetic}.
To overcome this generic issue in ES, recent advances have been proposed to improve the performance of ES algorithms. 
One such improvement is \textit{Policy Gradients with Parameter-Based Exploration (PGPE)}~\cite{sehnke2010parameter}, which estimates gradients in a black-box fashion so the computation of fitness does not have to be differentiable \emph{per se}.  
Since PGPE runs linear to the number of parameters for each iteration, it is an efficient and the go-to algorithm in many scenarios.
With the estimated gradients, gradient-based optimizers such as Adam~\cite{kingma2014adam} can be used for optimization, 
while there are also work such as ClipUp~\cite{toklu2020clipup} offering a simpler and more efficient optimizer specifically tailored for PGPE. 
Another representative ES algorithm is \textit{Covariance matrix adaptation evolution strategy (CMA-ES)}, which in practice is considered more performant than PGPE.
However, it runs in the quadratic time w.r.t. the number of parameters for each iteration, which limits its use in many problems with larger numbers of parameters where PGPE is still feasible.

\textbf{Language-derived Image Generation} has been seeing very recent trends in creativity setting, where there are several directions to leverage CLIP~\cite{radford2021learning}, a pre-trained model with two encoders, one for image and one for text, that can convert images and text into the same, comparable low-dimensional embedding space.
As the image encoder is a differentiable neural network, it can provide a gradient to the output of a differentiable generative model. 
The gradient can be further back-propagated through the said model till its parameters.
For example, one direction of works uses CLIP's gradient to guide a GAN's generator, such as guiding BigGAN~\cite{wang2021bigsleep}, guiding VQGAN~\cite{samburtonking2021introduction}, guiding Siren~\cite{wang2021deepdaze}, or a GAN with genetic algorithm-generated latent space~\cite{galatolo2021generating}. 
Another direction of work applies CLIP to differentiable renderers. 
CLIPDraw~\cite{frans2021clipdraw} proposes to generate the images with diffvg~\cite{li2020differentiable}, a differentiable SVG renderer. 
Although all these methods use the same pre-trained CLIP model for guidance, they show a drastically different artistic property, for which we hypothesize that the art style is determined by the intrinsic properties of ``painter'', i.e., the GAN generator or renderer.

%\section{Modern Evolution Strategies based Creativity}
\section{Modern Evolution Strategies for Creativity}
\label{sec:method}
\vskip -0.3cm
The architecture of our proposed pipeline is shown in Figure~\ref{fig:architecture}.
Our proposed method synthesizes painting by placing transparent triangles using evolution strategy (ES).
Overall, we can represent a configuration of triangles in a parameter space which composes of positions and colors of triangles,
render such configuration onto a canvas,
and calculate its fitness based on how well the rendered canvas fits a target image or an concept in the form of a text prompt.
The ES algorithm keeps a pool of candidate configurations and uses mutations to evolves better ones measured by the said fitness.
To have better creative results, we use a modern ES algorithm, PGPE~\cite{sehnke2010parameter} optimized by ClipUp~\cite{toklu2020clipup} optimizer. 
Engineering-wise we use the pgpelib~\cite{toklu2008pgpelib} implementation of PGPE and ClipUp.

As we choose to follow the spirit of minimalist art, we use transparent triangles as the parameter space.
Concretely, a configuration of $N$ triangles is parameterized by a collection of $(x_1, y_1, x_2, y_2, x_3, y_3, r,g,b,a)$ for each of the triangles,
which are vertex coordinates and the RGBA (Red, Green, Blue, and Alpha a.k.a. transparency channel) color, totally making $10N$ parameters.
In the ES, we update all parameters and use a fixed hyper-parameter, the number of triangles $N$.
Note that $N$ is better understood as the upper bound of number of triangles to use: although $N$ is fixed, the algorithm is still capable of effectively using ``fewer'' triangles by making unwanted ones transparent.

\begin{figure}[!htb]
\begin{small}
\captionsetup[subfigure]{labelformat=empty}
\centering
\vskip -0.7cm
\begin{subfigure}[h]{1.0\textwidth}
\newcommand{\imgwidth}{2.2cm}
\centering
\begin{tabular}{ccccc}
    \toprule
    \textbf{Target Image} & \textbf{10 Triangles} & \textbf{25 Triangles} & \textbf{50 Triangles} & \textbf{200 Triangles} \\
    \midrule
    \includegraphics[width=\imgwidth]{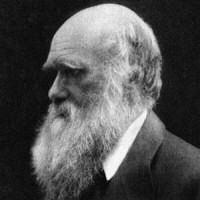} &
        \includegraphics[width=\imgwidth]{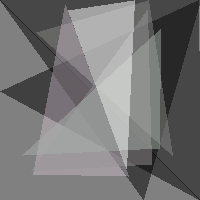} &
        \includegraphics[width=\imgwidth]{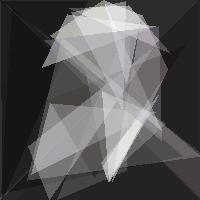} &
        \includegraphics[width=\imgwidth]{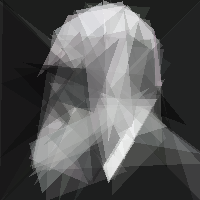} &
        \includegraphics[width=\imgwidth]{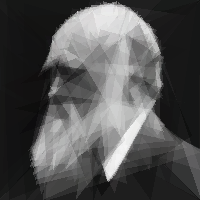} \\
    ``Darwin'' & Fitness = $96.82\%$ & Fitness = $99.25\%$ & Fitness = $99.51\%$ & Fitness = $99.75\%$ \\
    \midrule
    \includegraphics[width=\imgwidth]{es-bitmap-target-monalisa.png} &
        \includegraphics[width=\imgwidth]{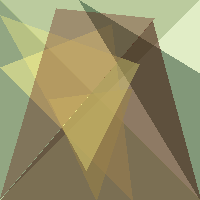} &
        \includegraphics[width=\imgwidth]{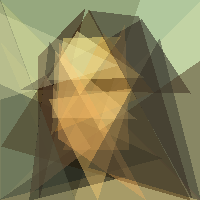} &
        \includegraphics[width=\imgwidth]{es-bitmap-fit-monalisa-50-run-1-coalesce-1x1.png} &
        \includegraphics[width=\imgwidth]{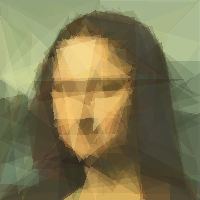} \\
    ``Mona Lisa'' & Fitness = $98.02\%$ & Fitness = $99.30\%$ & Fitness = $99.62\%$ & Fitness = $99.80\%$ \\
    \midrule
    \includegraphics[width=\imgwidth]{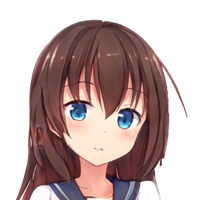} &
        \includegraphics[width=\imgwidth]{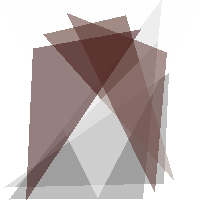} &
        \includegraphics[width=\imgwidth]{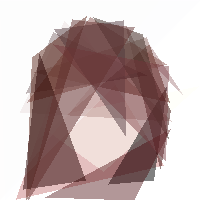} &
        \includegraphics[width=\imgwidth]{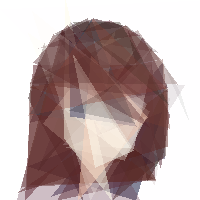} &
        \includegraphics[width=\imgwidth]{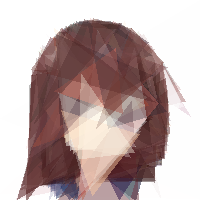} \\
    ``Anime Face'' & Fitness = $94.97\%$ & Fitness = $98.17\%$ & Fitness = $98.80\%$ & Fitness = $99.07\%$ \\
    \midrule
    \includegraphics[width=\imgwidth]{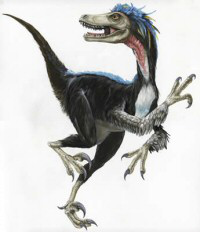} &
        \includegraphics[width=\imgwidth]{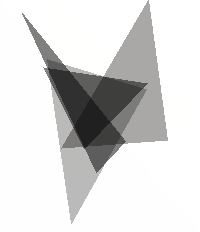} &
        \includegraphics[width=\imgwidth]{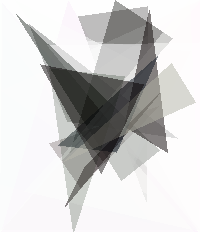} &
        \includegraphics[width=\imgwidth]{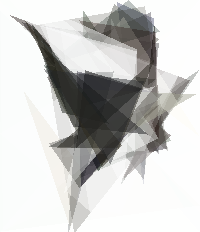} &
        \includegraphics[width=\imgwidth]{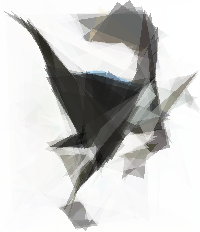} \\
    ``Velociraptor'' & Fitness = $95.76\%$ & Fitness = $98.03\%$ & Fitness = $99.00\%$ & Fitness = $99.25\%$ \\
    \midrule
    \includegraphics[width=\imgwidth]{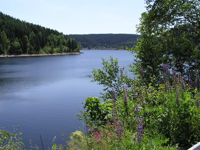} &
        \includegraphics[width=\imgwidth]{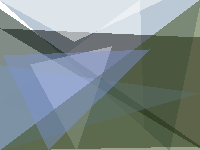} &
        \includegraphics[width=\imgwidth]{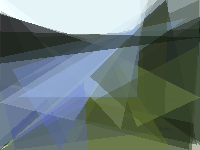} &
        \includegraphics[width=\imgwidth]{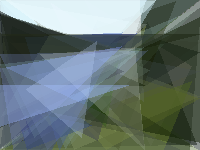} &
        \includegraphics[width=\imgwidth]{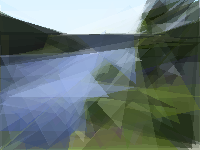} \\
    ``Landscape'' & Fitness = $97.07\%$ & Fitness = $98.83\%$ & Fitness = $99.08\%$ & Fitness = $99.25\%$ \\
    \midrule
    \includegraphics[width=\imgwidth]{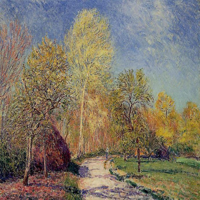} &
        \includegraphics[width=\imgwidth]{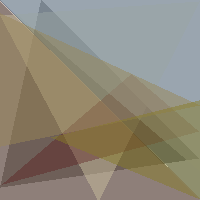} &
        \includegraphics[width=\imgwidth]{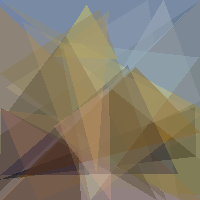} &
        \includegraphics[width=\imgwidth]{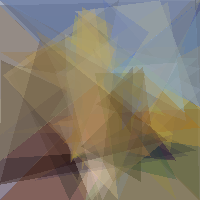} &
        \includegraphics[width=\imgwidth]{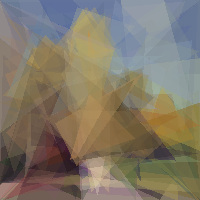} \\
    ``Impressionism'' & Fitness = $98.82\%$ & Fitness = $99.23\%$ & Fitness = $99.34\%$ & Fitness = $99.48\%$ \\
    \bottomrule
\end{tabular}
\end{subfigure}
\vskip -0.1cm
\caption{Qualitative and quantitative results from fitting several targets with $10$, $25$, $50$, and $200$ triangles, each running for 10,000 steps.
Images credits: Darwin, Mona Lisa, Velociraptor are from \cite{alteredqualia2008evolutiongenetic}. Anime Face is generated by Waifu Labs \cite{sizigi2019waifu}. Landscape is from Wikipedia~\cite{wiki:landscape}. Impressionism is \textit{A May Morning in Moret} by Alfred Sisley, collected by \cite{gonsalves2021impressionist}.
Fitness is $100\% - $ L2 Loss.
}
\vskip -0.7cm
\label{fig:es-ours-different-n-triangles}
\end{small}
\end{figure}

As the ES is orthogonal to the concrete fitness evaluation, we are left with many free choices regarding what counts as fitting.
Particularly, we consider two kinds of fitness, namely, fitting a concrete image (the lower branch in Figure~\ref{fig:architecture}) and fitting a concept (the upper branch in Figure~\ref{fig:architecture}).
Fitting a concrete image is straightforward, where we can simply use the pixel-wise L2 loss between the rendered canvas and the target image as the fitness.
Fitting a concept requires more elaboration. We represent the concept as a text prompt and embed the text prompt using the text encoder in CLIP~\cite{radford2021learning} which we discuss in detail in Section~\ref{sec:realted-work}.
Then we embed the rendered canvas using the image encoder also available in CLIP. Since the CLIP models are trained so that both embedded images and texts are comparable under Cosine distance for similarity, we use such distance as the fitness.
We note that since the ES algorithm provides black-box optimization, the renderer, like fitness computation, does not necessarily need to be differentiable.

We find in practice a few decisions should be made so the whole pipeline can work reasonably well. 
First, we augment the rendered canvas by random cropping in calculating the fitness and average the fitness on each of the augmented canvas, following the practice of \cite{samburtonking2021introduction,frans2021clipdraw}. 
This would prevent the rendered canvas from overfitting and increase the stability in the optimization.
Second, we render the triangles on top of a background with a uniform distribution noise. Mathematically, this equals to modeling the uncertainty of parts in the canvas not covered by triangles with a max-entropy assumption, and using Monte Carlo method for approximation.
Finally, we limit the maximal alpha value for each triangle to $0.1$, which prevents front triangles from (overly) shadowing the back ones.

\section{Fitting Concrete Target Image}
\vskip -0.1cm
In this section, we show the performance of our proposed work on fitting a concrete target image.
In doing so, the model takes the lower branch in Figure~\ref{fig:architecture}.
We fit the famous painting ``Mona Lisa'' with $50$ triangles by running evolution for $10,000$ steps in Figure~\ref{fig:es-ours-mona-lisa}. 
Our result is a distinctive art style represented by well-placed triangles that care both fine-grained textures and large backgrounds. 
The evolution process also displays the coarse-to-fine adjustments of the shapes' positions and colors.

\begin{figure}[!htb]
\captionsetup[subfigure]{labelformat=empty}
\centering
\vskip -0.1cm
\begin{subfigure}[h]{1.0\textwidth}
    \newcommand{\imgwidth}{2.3cm}
    \centering
    \begin{tabular}{cccc}
    \toprule
    \textbf{Target Image} &
        \textbf{
            \begin{tabular}{@{}c@{}}Ours \\ (10,000 iteration)\end{tabular}
        } & 
        \textbf{
            \begin{tabular}{@{}c@{}}Basic\\ (10,000 iteration)\end{tabular}
        } & 
        \textbf{
            \begin{tabular}{@{}c@{}}Basic\\ (560,000 iteration)\end{tabular}
        }\\
    \includegraphics[width=\imgwidth]{es-bitmap-target-monalisa.png} &
        \includegraphics[width=\imgwidth]{es-bitmap-fit-monalisa-50-run-1-coalesce-1x1.png} &
        \includegraphics[width=\imgwidth]{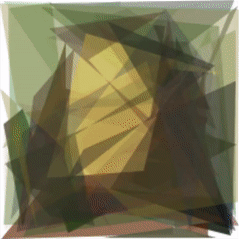} &
        \includegraphics[width=\imgwidth]{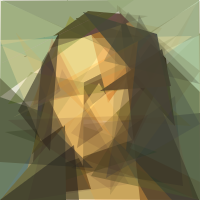} \\
    ``Mona Lisa'' & \textbf{Fitness = $\mathbf{99.62\%}$} & Fitness = $97.23\%$ & Fitness = $99.58\%$ \\
    \bottomrule
    \end{tabular}
\end{subfigure}
\vskip 0.1cm 
\begin{subfigure}[h]{1.0\textwidth}
    \newcommand{\imgwidth}{12cm}
    \centering
    \begin{tabular}{p{\imgwidth}}
        \includegraphics[width=\imgwidth]{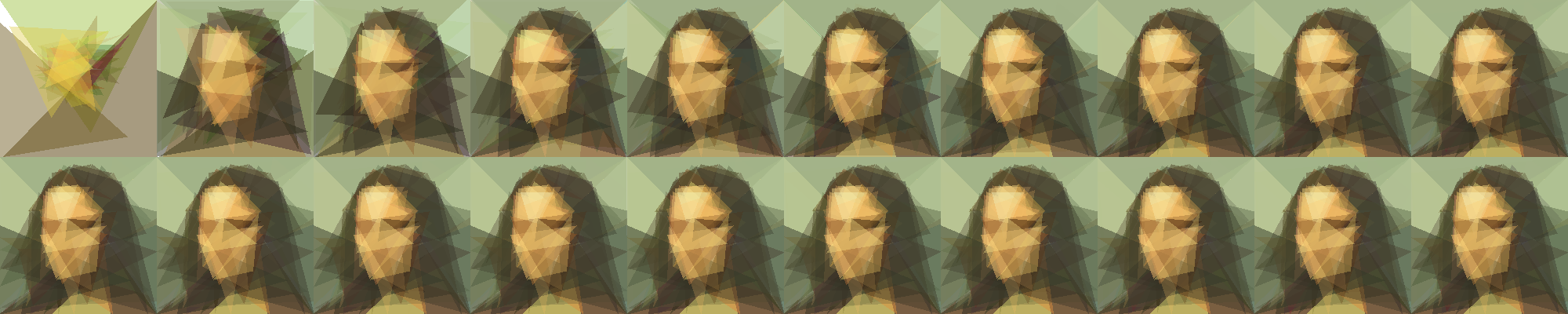} \\
        \midrule
        \includegraphics[width=\imgwidth]{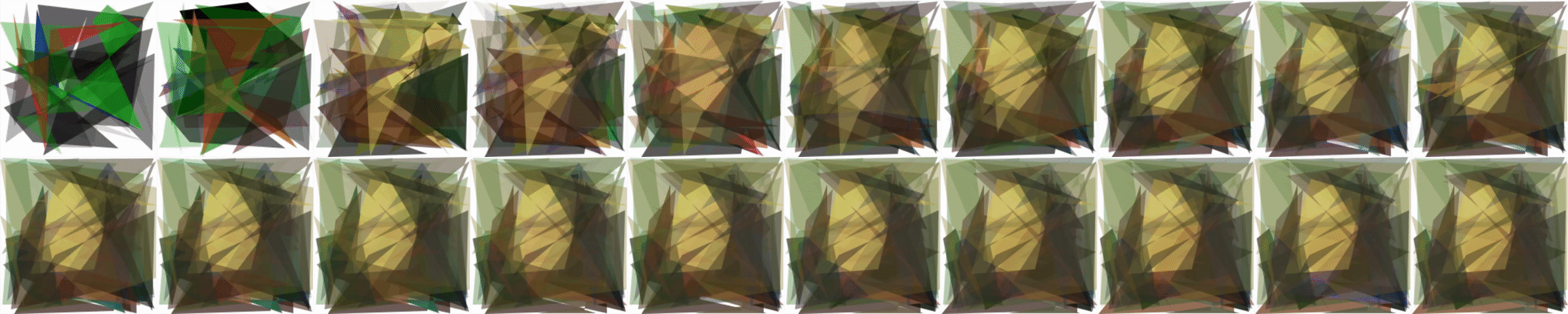} \\
    \bottomrule
    \end{tabular}
\end{subfigure}
\vskip -0.015cm
\caption{Compare choices of evolution algorithm: Ours (PGPE with ClipUp) vs.\ basic evolution algorithm (mutation with simulated annealing)~\cite{alteredqualia2008evolutiongenetic}. Both settings fit 50 triangles and all choices except for the EA are the same. 
We show the results of ours and the basic algorithm at the end of $10,000$ iterations, and the result of the basic algorithm after running $56$ times more iterations.
The details of evolution process until $10,000$ iterations are shown in the bottom half, where the upper group is for ours and the lower group is for the basic algorithm. Fitness is $100\% - $ L2 Loss.
}
\vskip -0.5cm
\label{fig:es-ours-vs-basic}
\end{figure}

\textbf{Number of triangles and parameters}.
Our proposed pipeline is able to fit any target images and could handle a wide range of number of parameters, since PGPE runs efficiently, i.e., linear to the number of parameters. 
This is demonstrated by applying our method to fit several target images with $10$, $25$, $50$, $200$ triangles, which corresponds to $100$, $250$, $500$ and $2000$ parameters respectively.
As shown in Figure~\ref{fig:es-ours-different-n-triangles},
our proposed pipeline works well for a wide range of target images, and the ES algorithm is capable of using the number of triangles as a ``computational budget'' where extra triangles could always be utilized for gaining in fitness. 
This allows a human artist to use the number of triangles in order to find the right balance between abstractness and details in the produced art.

\textbf{Choice of ES Algorithm}.
We compare two choices of evolution algorithm: ours, which uses the recent PGPE with ClipUp, and a basic, traditional one, which consists of mutation and simulated annealing adopted earlier~\cite{johansson2008genetic,alteredqualia2008evolutiongenetic}. 
As shown in Figure~\ref{fig:es-ours-vs-basic}, our choice of more recent algorithms leads to better results than the basic one under the same parameter budget.
Subjectively, our final results are more visually closer to the target image with a smoother evolution process, and quantitatively, our method leads to much better fitness ($99.62\%$ vs. $97.23\%$). 
Furthermore, even allowing $56$ times more iterations for the basic algorithm does not lead to results better than ours.

\begin{figure}[!htb]
\captionsetup[subfigure]{labelformat=empty}
\centering
\vskip -0.3cm
\begin{subfigure}[h]{1.0\textwidth}
    \newcommand{\imgwidth}{3.0cm}
    \centering
    \begin{tabular}{ccc}
    \toprule
    \textbf{Target Image} &
        \textbf{
            \begin{tabular}{@{}c@{}}Evolution Strategy \\ (Non-gradient)\end{tabular}
        } & 
        \textbf{
            \begin{tabular}{@{}c@{}}Differentiable Renderer \\ (Gradient-based)\end{tabular}
        } \\
    \includegraphics[width=\imgwidth]{es-bitmap-target-monalisa.png} &
        \includegraphics[width=\imgwidth]{es-bitmap-fit-monalisa-200-run-1-coalesce-1x1.png} &
        \includegraphics[width=\imgwidth]{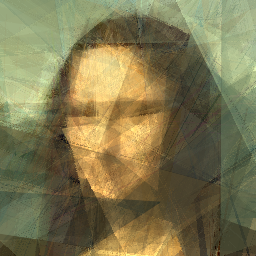} \\
    ``Mona Lisa'' & Fitness = $\mathbf{99.80\%}$ & Fitness = $99.68\%$ \\
    \bottomrule
    \end{tabular}
\end{subfigure}
\vspace{0.1cm}
\begin{subfigure}[h]{1.0\textwidth}
    \newcommand{\imgwidth}{12cm}
    \centering
    \begin{tabular}{p{\imgwidth}}
        \includegraphics[width=\imgwidth]{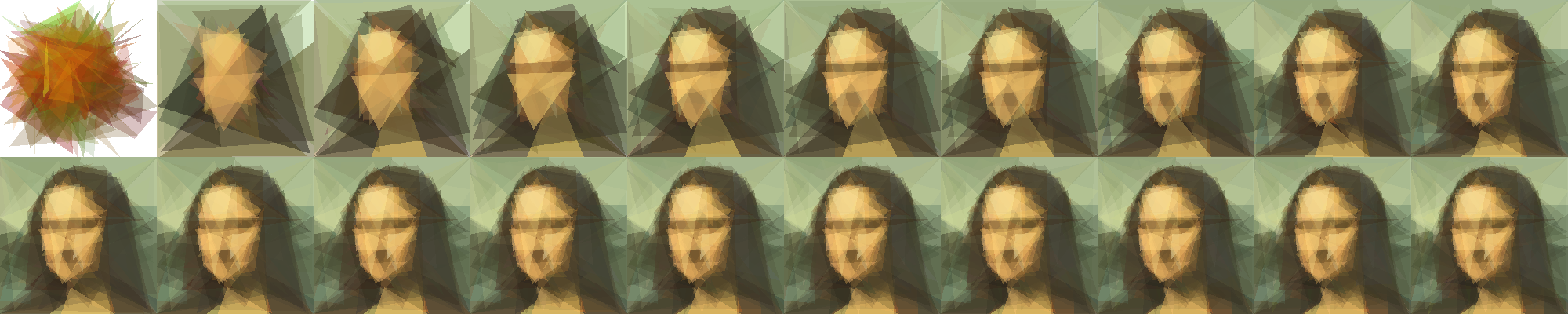} \\
        \midrule
        \includegraphics[width=\imgwidth]{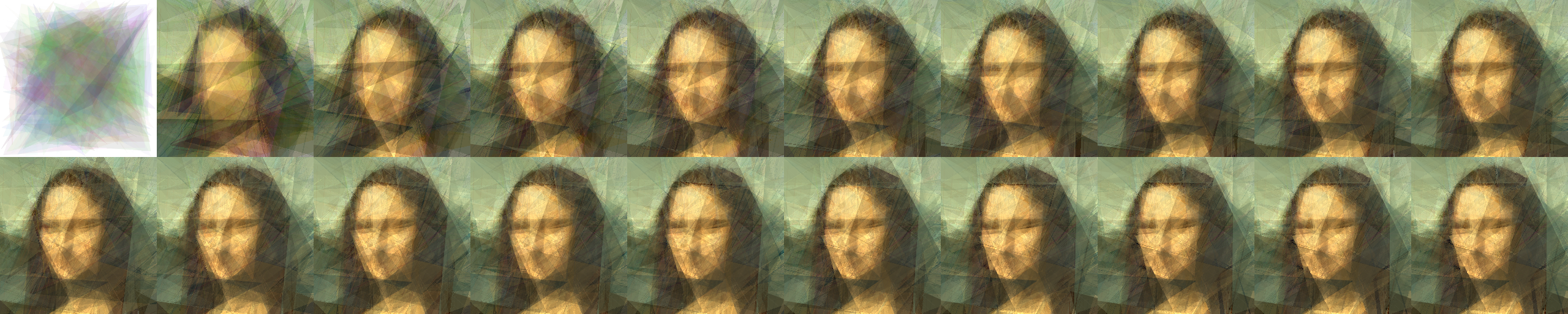} \\
    \bottomrule
    \end{tabular}
\end{subfigure}
\vskip -0.1cm
\caption{Evolution strategies (non-gradient method) vs. differentiable renderer (gradient based method) fitting target image with $200$ triangles. The upper half shows the final results and the bottom half shows the details of optimization. Fitness is $100\% - $ L2 Loss.}
\label{fig:es-vs-diff}
\vskip -0.4cm
\end{figure}

\textbf{Comparison with Gradient-based Optimization}.
While our proposed approach is ES-based, it is interesting to investigate how it compares to gradient-based optimization since the latter is commonly adopted recently (See Section~\ref{sec:realted-work}).
Therefore we conduct a gradient-based setup by implementing rendering of composed triangles using nvdiffrast~\cite{laine2020modular}, a point-sampling-based differentiable renderer.
We use the same processing as mentioned in Section~\ref{sec:method}.
As shown in Figure~\ref{fig:es-vs-diff}, our proposed ES-based method can achieve similar yet slightly higher fitness than results compared with the gradient-optimized differentiable renderer.
Furthermore and perhaps more interestingly, two methods produce artworks with different styles: 
our proposed method can adaptive allocating large triangles for background and small ones for detailed textures, 
whereas the differentiable renderer tends to introduce textures unseen in the target image (especially in the background). 
We argue that due to the difference in the optimization mechanism, our method focuses more on the placement of triangles while the differentiable renderer pays attention to the compositing of transparent colors.

\begin{figure}[!htb]
\captionsetup[subfigure]{labelformat=empty}
\centering
\vskip -0.3cm
\begin{subfigure}[h]{1.0\textwidth}
\newcommand{\txtwidth}{2cm}
\newcommand{\imgwidth}{2.7cm}
\newcommand{\procwidth}{8.3cm}
% "\raisebox{-.5\height}{ ... }" below is for vertically align images with text in tabular.
\centering
\begin{tabular}{m{\txtwidth}cc}
    \toprule
    \textbf{Prompt} & \textbf{Evolved Results} & \textbf{Evolution Process} \\
    \midrule
    ``Self''
     & 
        \raisebox{-.5\height}{\includegraphics[width=\imgwidth]{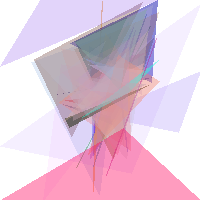}} &
        \raisebox{-.5\height}{\includegraphics[width=\procwidth]{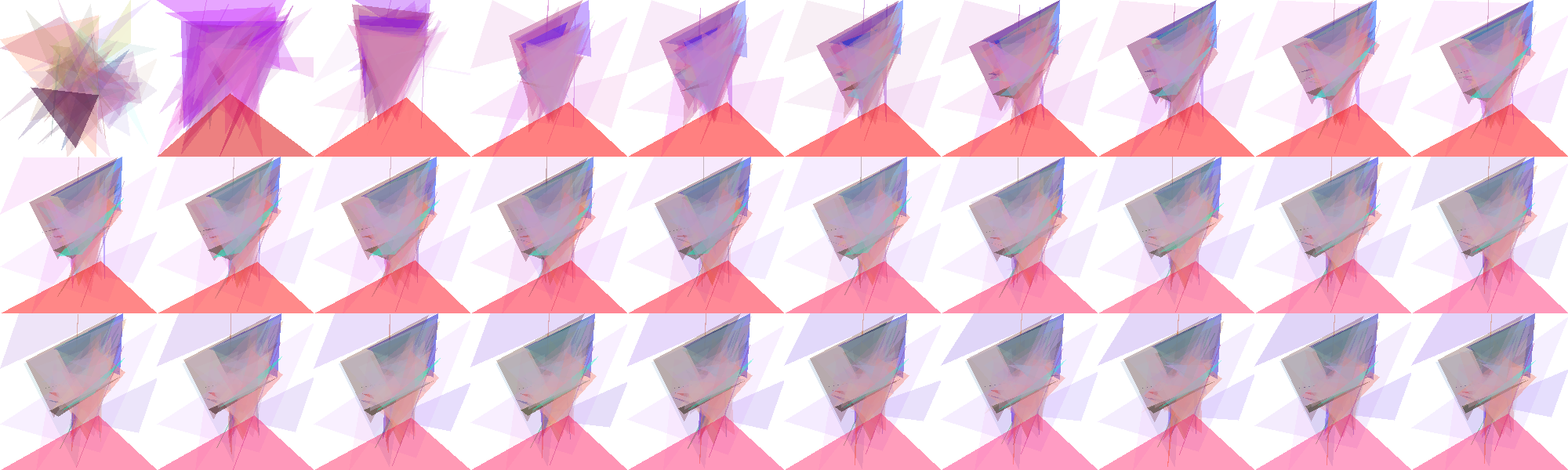}} \\
    \midrule
    ``Human''
     & 
        \raisebox{-.5\height}{\includegraphics[width=\imgwidth]{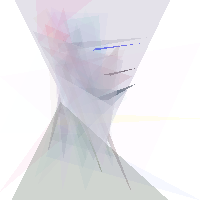}} &
        \raisebox{-.5\height}{\includegraphics[width=\procwidth]{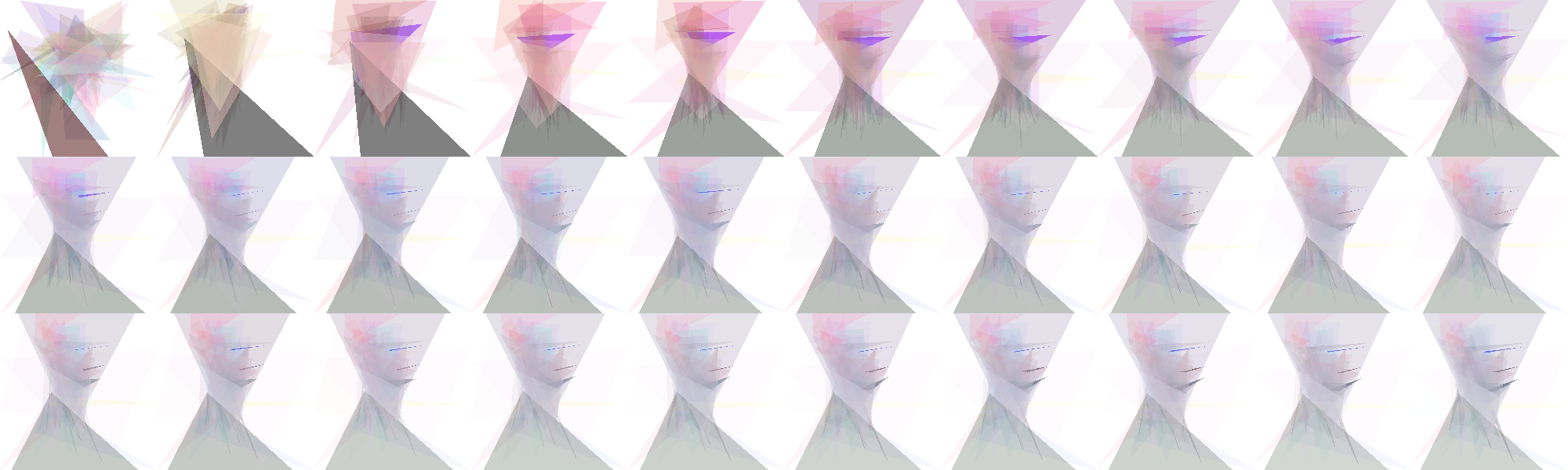}} \\
    \midrule
    ``Walt Disney World''
     & 
        \raisebox{-.5\height}{\includegraphics[width=\imgwidth]{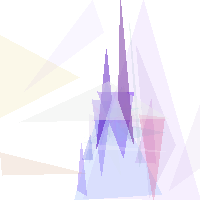}} &
        \raisebox{-.5\height}{\includegraphics[width=\procwidth]{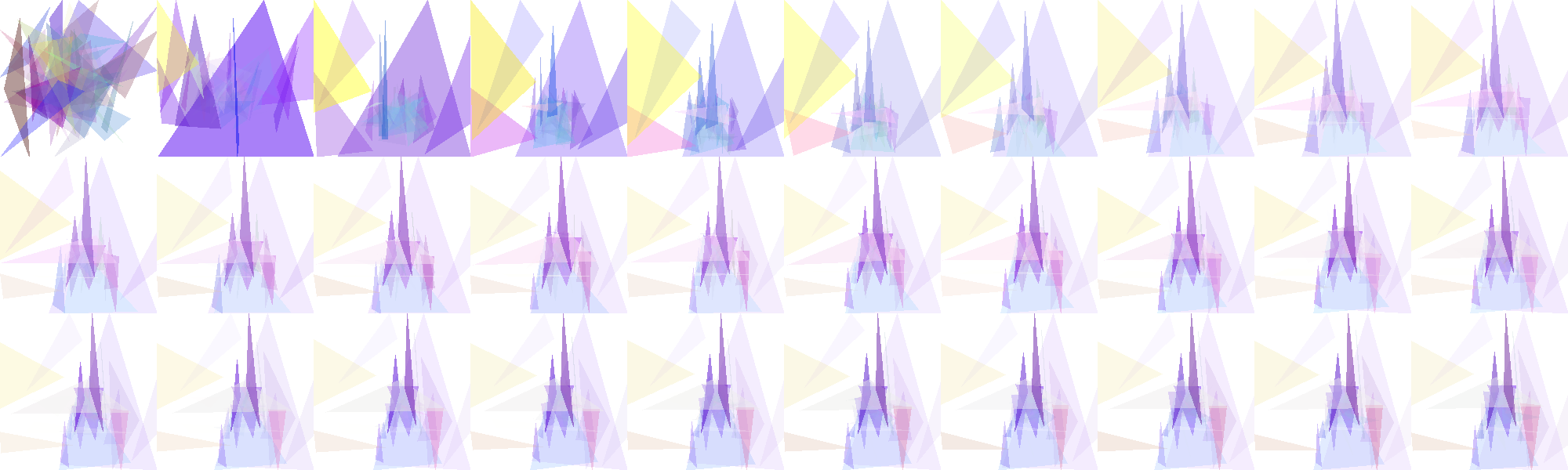}} \\
    \midrule
    {\tiny ``The corporate headquarters complex of Google located at 1600 Amphitheatre Parkway in Mountain View, California.''}
     & 
        \raisebox{-.5\height}{\includegraphics[width=\imgwidth]{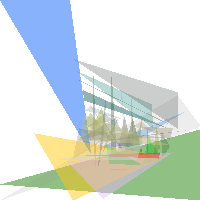}} & 
        \raisebox{-.5\height}{\includegraphics[width=\procwidth]{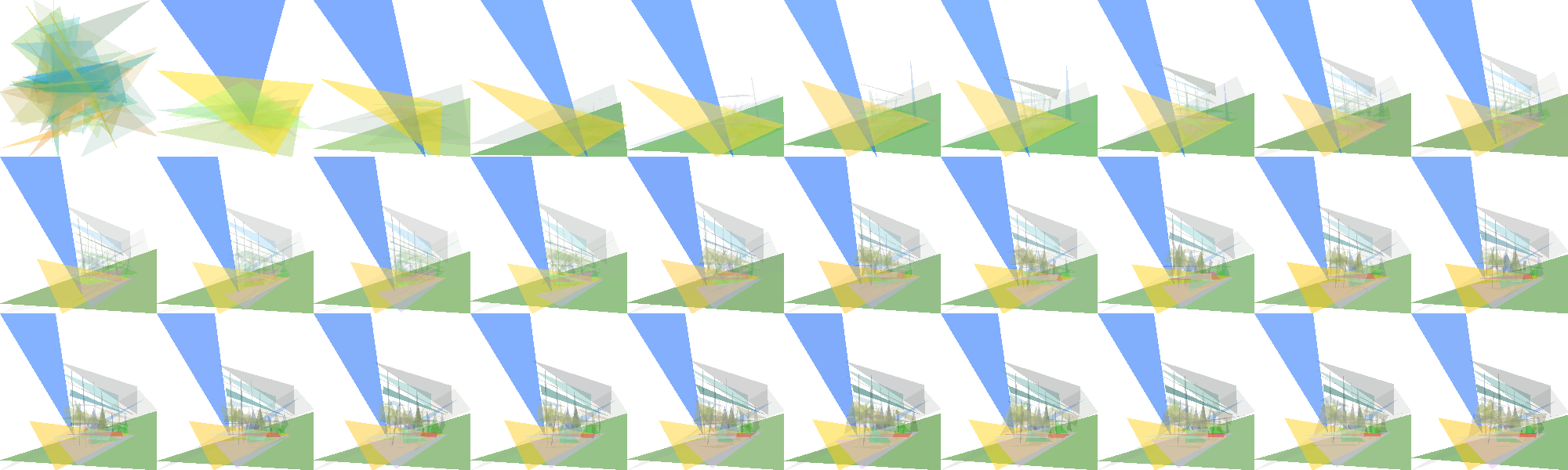}} \\
    \bottomrule
\end{tabular}
\end{subfigure}
\vskip -0.1cm
\caption{ES and CLIP fit the concept represented in text prompt, using 50 triangles and running evolution for $2,000$ steps. 
Each row shows the text prompt followed by the finally evolved results and the evolution process. 
We show exemplary prompts for 3 kinds of text, ranging from a single word (``Self'' and ``Human''), a phrase (``Walt Disney Land''), and a long sentence (``The corporate headquarters complex of Google located at 1600 Amphitheatre Parkway in Mountain View, California.'').
}
\label{fig:es-clip-examples}
\vskip -0.4cm
\end{figure}

\section{Fitting Abstract Concept with CLIP}
\vskip -0.3cm
In this section, we show the performance of our method configured to fit an abstract concept represented by language. 
In doing so, the model takes the upper branch in Figure~\ref{fig:architecture}.
Formally, the parameter space remains the same, but the fitness is calculated as the cosine distance between the text prompt and the rendered canvas, both encoded by CLIP. 
Since the model is given more freedom to decide what to paint, this problem is arguably a much harder yet more interesting problem than fitting concrete images in the previous section.

In Figure~\ref{fig:es-clip-examples}, we show the evolution result and process of fitting abstract concept represented as text prompt, using 50 triangles and running evolution for $2,000$ steps. 
We found that unlike fitting a concrete images, $2,000$ steps is enough for fitting a concept to converge.  
Our method could handle text prompts ranging from a single word to a phrase, and finally, to a long sentence, even though the task itself is arguably more challenging than the previous one.
The results show a creative art concept that is abstract, not resembling a particular image, yet correlated with humans' interpretation of the text.
The evolution process also demonstrates iterative adjustment, such as the human shape in the first two examples, the shape of castles in Disney World, as well as in the final example, the cooperate-themed headquarters.
Also, compared to fitting concrete images in the previous section, our method cares more about the placement of triangles.

\begin{figure}[!htb]
\captionsetup[subfigure]{labelformat=empty}
\centering
\vskip -0.3cm
\newcommand{\txtwidth}{2cm}
\newcommand{\imgwidth}{2.3cm}
% "\raisebox{-.5\height}{ ... }" below is for vertically align images with text in tabular.
\centering
\begin{tabular}{m{\txtwidth}cccc}
    \toprule
    \textbf{Prompt} & \textbf{10 Triangles} & \textbf{25 Triangles} & \textbf{50 Triangles} & \textbf{200 Triangles} \\
    \midrule
    ``Self'' & 
        \raisebox{-.5\height}{\includegraphics[width=\imgwidth]{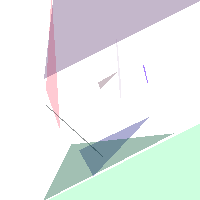}} &
        \raisebox{-.5\height}{\includegraphics[width=\imgwidth]{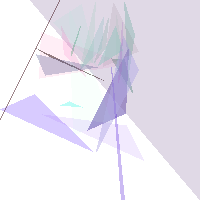}} &
        \raisebox{-.5\height}{\includegraphics[width=\imgwidth]{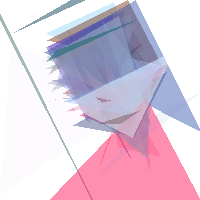}} &
        \raisebox{-.5\height}{\includegraphics[width=\imgwidth]{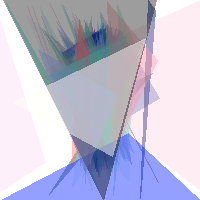}} \\
    \midrule
    ``Human'' & 
        \raisebox{-.5\height}{\includegraphics[width=\imgwidth]{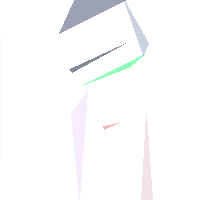}} &
        \raisebox{-.5\height}{\includegraphics[width=\imgwidth]{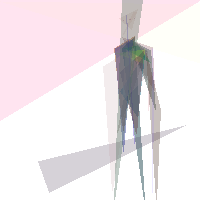}} &
        \raisebox{-.5\height}{\includegraphics[width=\imgwidth]{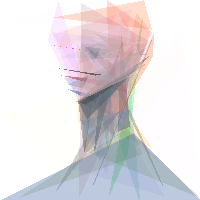}} &
        \raisebox{-.5\height}{\includegraphics[width=\imgwidth]{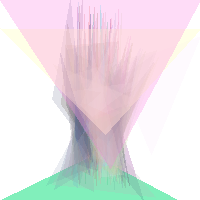}} \\
    \midrule
    ``Walt Disney World'' & 
        \raisebox{-.5\height}{\includegraphics[width=\imgwidth]{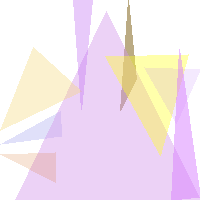}} &
        \raisebox{-.5\height}{\includegraphics[width=\imgwidth]{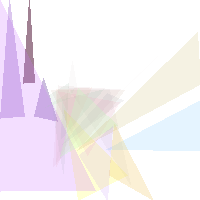}} &
        \raisebox{-.5\height}{\includegraphics[width=\imgwidth]{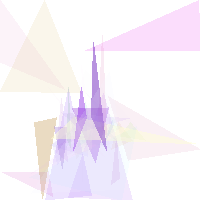}} &
        \raisebox{-.5\height}{\includegraphics[width=\imgwidth]{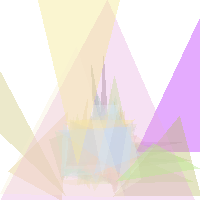}} \\
    \midrule
    ``A picture of Tokyo'' & 
        \raisebox{-.5\height}{\includegraphics[width=\imgwidth]{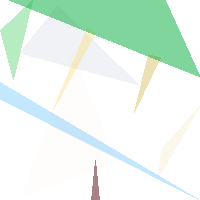}} &
        \raisebox{-.5\height}{\includegraphics[width=\imgwidth]{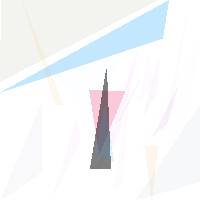}} &
        \raisebox{-.5\height}{\includegraphics[width=\imgwidth]{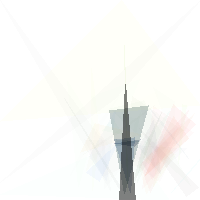}} &
        \raisebox{-.5\height}{\includegraphics[width=\imgwidth]{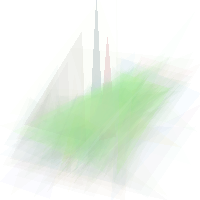}} \\
    \midrule
    {\tiny ``The corporate head\-quarters complex of Google located at 1600 Amphitheatre Parkway in Mountain View, California.''} & 
        \raisebox{-.5\height}{\includegraphics[width=\imgwidth]{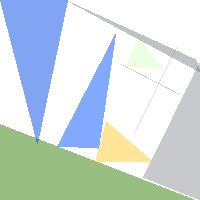}} &
        \raisebox{-.5\height}{\includegraphics[width=\imgwidth]{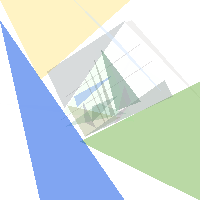}} &
        \raisebox{-.5\height}{\includegraphics[width=\imgwidth]{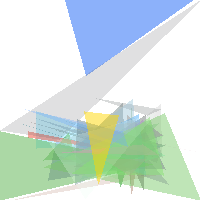}} &
        \raisebox{-.5\height}{\includegraphics[width=\imgwidth]{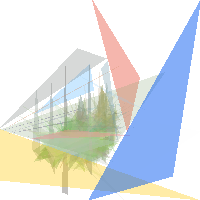}} \\
    \midrule
    {\tiny ``The United States of America commonly known as the United States or America is a country primarily located in North America.''} & 
        \raisebox{-.5\height}{\includegraphics[width=\imgwidth]{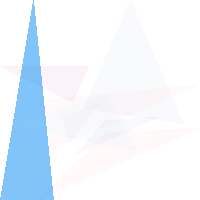}} &
        \raisebox{-.5\height}{\includegraphics[width=\imgwidth]{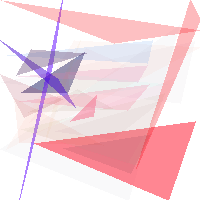}} &
        \raisebox{-.5\height}{\includegraphics[width=\imgwidth]{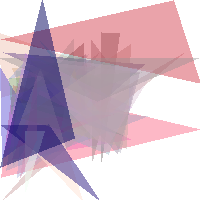}} &
        \raisebox{-.5\height}{\includegraphics[width=\imgwidth]{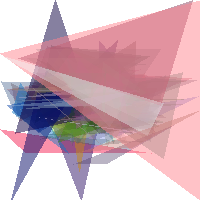}} \\
    \bottomrule
\end{tabular}
\vskip -0.1cm
\caption{Qualitative results from ES and CLIP fitting several text prompt with 10,25,50, and 200 triangles, each running for 2,000 steps. We show exemplary prompts for 3 kinds of text, ranging from a single word (``Self'' and ``Human''), to phrases (``A picture of Tokyo''), to long sentences.}
\label{fig:es-clip-different-n-triangles}
\vskip -0.7cm
\end{figure}

\textbf{Number of triangles and parameters}.
Like fitting a concrete image, we can also fit an abstract concept with a wide range of number of parameters since the PGPE algorithm and the way we represent canvas remains the same.
In Figure~\ref{fig:es-clip-different-n-triangles} we apply our method to fit several concept (text prompt) with $10$, $25$, $50$, $200$ triangles, which corresponds to $100$, $250$, $500$ and $2000$ parameters respectively.
It is shown that our proposed pipeline is capable of leveraging the number of triangles as a ``budget for fitting'' to balance between the details and the level of abstraction. 
Like in the previous task, this allows a human artist to balance the abstractness in the produced art.

We observe that while the model could comfortably handle at least up to $50$ triangles, more triangles ($200$) sometimes poses challenges: For example, with $200$ triangles, ``corporate headquarters \textellipsis'' gets a better result while ``a picture of Tokyo'' leads to a poor one. This may be due to the difficulties composing overly shadowed triangles, and we leave it for future study. 

\begin{figure}[!htb]
\begin{small}
\captionsetup[subfigure]{labelformat=empty}
\centering
\newcommand{\txtwidth}{2cm}
\newcommand{\imgwidth}{2.5cm}
% "\raisebox{-.5\height}{ ... }" below is for vertically align images with text in tabular.
\centering
\begin{tabular}{m{\txtwidth}cccc}
    \toprule
    \textbf{Prompt} & \multicolumn{4}{c}{\textbf{4 Individual Runs}} \\
    \midrule
    ``Self'' & 
        \raisebox{-.5\height}{\includegraphics[width=\imgwidth]{images/es-clip-_Self..._-50-run-2-1-coalesce-1x1.png}} &
        \raisebox{-.5\height}{\includegraphics[width=\imgwidth]{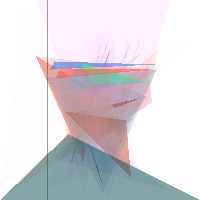}} &
        \raisebox{-.5\height}{\includegraphics[width=\imgwidth]{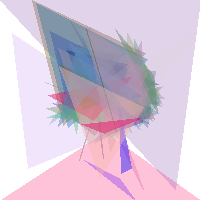}} &
        \raisebox{-.5\height}{\includegraphics[width=\imgwidth]{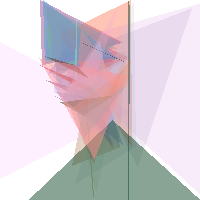}} \\
    \midrule
    ``Human'' & 
        \raisebox{-.5\height}{\includegraphics[width=\imgwidth]{images/es-clip-_Human..._-50-run-2-1-coalesce-1x1.png}} & 
        \raisebox{-.5\height}{\includegraphics[width=\imgwidth]{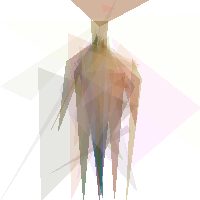}} & 
        \raisebox{-.5\height}{\includegraphics[width=\imgwidth]{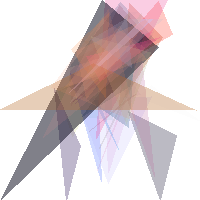}} & 
        \raisebox{-.5\height}{\includegraphics[width=\imgwidth]{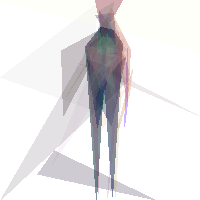}} \\
    \midrule
    ``Walt Disney World'' & 
        \raisebox{-.5\height}{\includegraphics[width=\imgwidth]{images/es-clip-_Walt_Disney_World..._-50-run-2-1-coalesce-1x1.png}} & 
        \raisebox{-.5\height}{\includegraphics[width=\imgwidth]{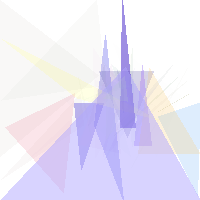}} & 
        \raisebox{-.5\height}{\includegraphics[width=\imgwidth]{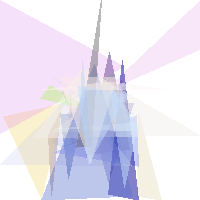}} & 
        \raisebox{-.5\height}{\includegraphics[width=\imgwidth]{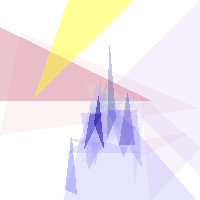}} \\
    \midrule
    ``A picture of Tokyo'' & 
        \raisebox{-.5\height}{\includegraphics[width=\imgwidth]{images/es-clip-_A_picture_of_Tokyo..._-50-run-2-1-coalesce-1x1.png}} & 
        \raisebox{-.5\height}{\includegraphics[width=\imgwidth]{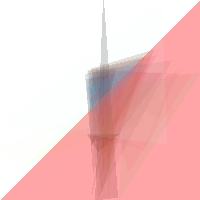}} & 
        \raisebox{-.5\height}{\includegraphics[width=\imgwidth]{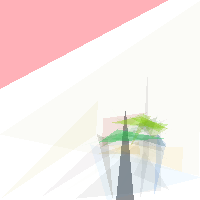}} & 
        \raisebox{-.5\height}{\includegraphics[width=\imgwidth]{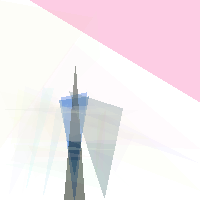}} \\
    \midrule
    {\tiny ``The corporate head\-quarters complex of Google located at 1600 Amphitheatre Parkway in Mountain View, California.''} & 
        \raisebox{-.5\height}{\includegraphics[width=\imgwidth]{images/es-clip-_The_corporate_headquarters_complex_of_Go..._-50-run-2-1-coalesce-1x1.png}} & 
        \raisebox{-.5\height}{\includegraphics[width=\imgwidth]{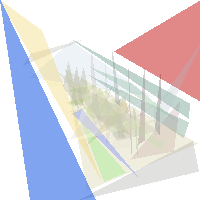}} & 
        \raisebox{-.5\height}{\includegraphics[width=\imgwidth]{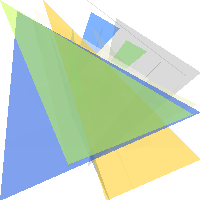}} & 
        \raisebox{-.5\height}{\includegraphics[width=\imgwidth]{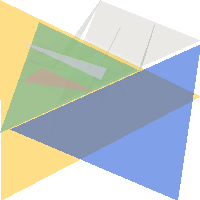}} \\
    \midrule
    {\tiny ``The United States of America commonly known as the United States or America is a country primarily located in North America.''} &
        \raisebox{-.5\height}{\includegraphics[width=\imgwidth]{images/es-clip-_The_United_States_of_America_commonly_kn..._-50-run-2-1-coalesce-1x1.png}} &
        \raisebox{-.5\height}{\includegraphics[width=\imgwidth]{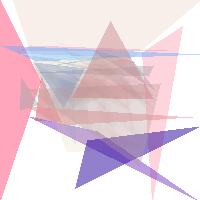}} &
        \raisebox{-.5\height}{\includegraphics[width=\imgwidth]{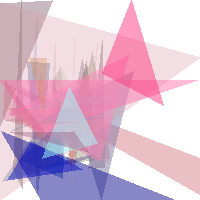}} &
        \raisebox{-.5\height}{\includegraphics[width=\imgwidth]{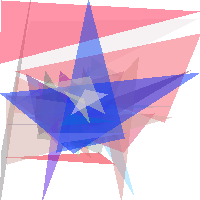}}  \\
    \bottomrule
\end{tabular}
\caption{Qualitative results from ES and CLIP fitting several text prompt with 50 triangles, each running for 2,000 steps. The text prompts follow that of Figure~\ref{fig:es-clip-different-n-triangles}. Results from four runs are shown.}
\label{fig:es-clip-multiple-runs}
\end{small}
\vskip -0.9cm
\end{figure}

\begin{figure}[!htb]
\captionsetup[subfigure]{labelformat=empty}
\centering
\vskip -0.3cm
\begin{subfigure}[h]{1.0\textwidth}
    \newcommand{\txtwidth}{3cm}
    \newcommand{\imgwidth}{3.0cm}
    \centering
    \begin{tabular}{m{\txtwidth}cc}
    \toprule
    \textbf{Prompt} &
        \textbf{
            \begin{tabular}{@{}c@{}}Evolution Strategy \\ (Non-gradient)\end{tabular}
        } & 
        \textbf{
            \begin{tabular}{@{}c@{}}Differentiable Renderer \\ (Gradient-based)\end{tabular}
        } \\
    \midrule
    ``Self'' &
        \raisebox{-.5\height}{\includegraphics[width=\imgwidth]{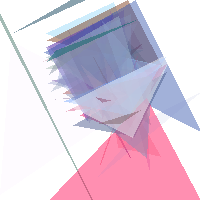}} &
        \raisebox{-.5\height}{\includegraphics[width=\imgwidth]{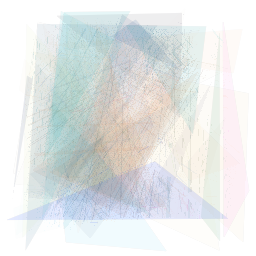}} \\
    \midrule
    ``Walt Disney World'' &
        \raisebox{-.5\height}{\includegraphics[width=\imgwidth]{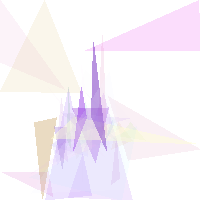}} &
        \raisebox{-.5\height}{\includegraphics[width=\imgwidth]{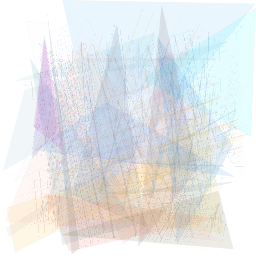}} \\
    \end{tabular}
\end{subfigure}
\vspace{0.1cm}
\begin{subfigure}[h]{1.0\textwidth}
    \newcommand{\imgwidth}{12cm}
    \centering
    \begin{tabular}{p{\imgwidth}}
        \toprule
        \includegraphics[width=\imgwidth]{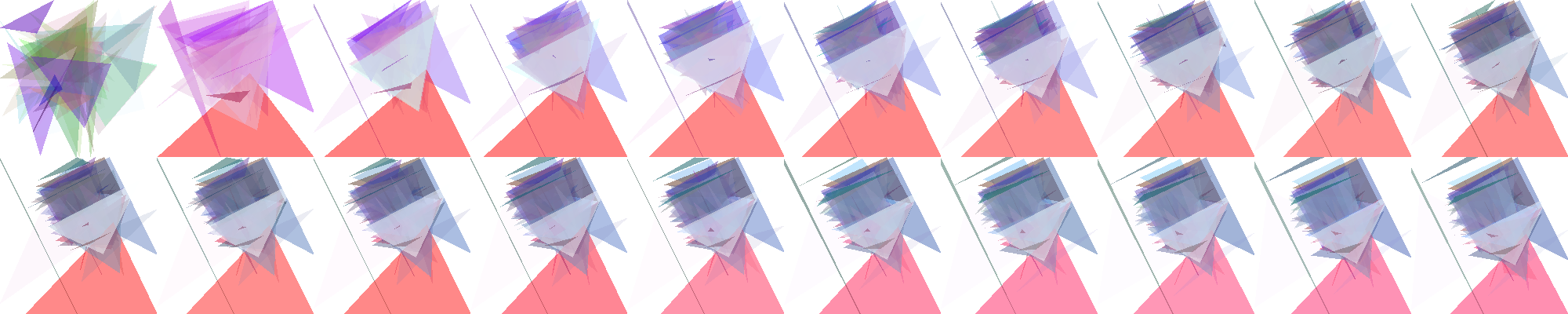} \\
        \midrule
        \includegraphics[width=\imgwidth]{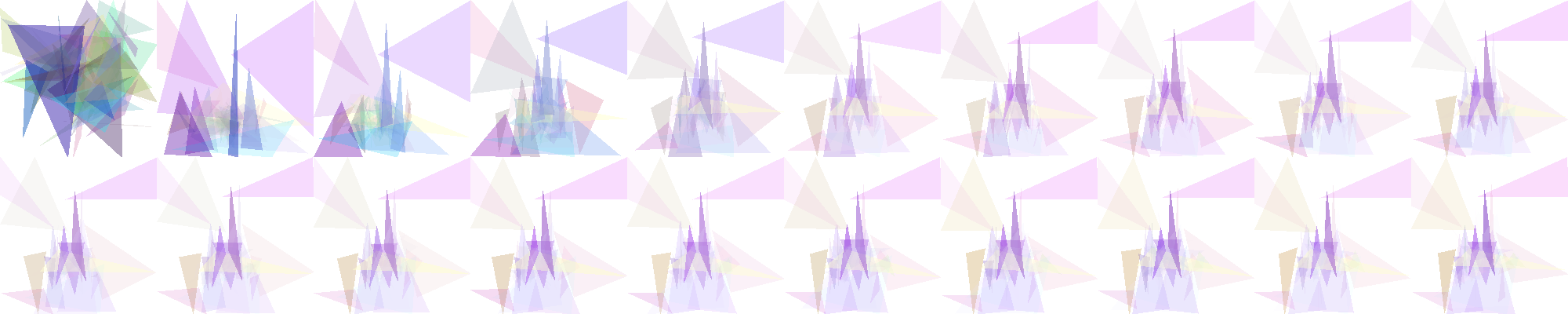} \\
        \midrule
        \includegraphics[width=\imgwidth]{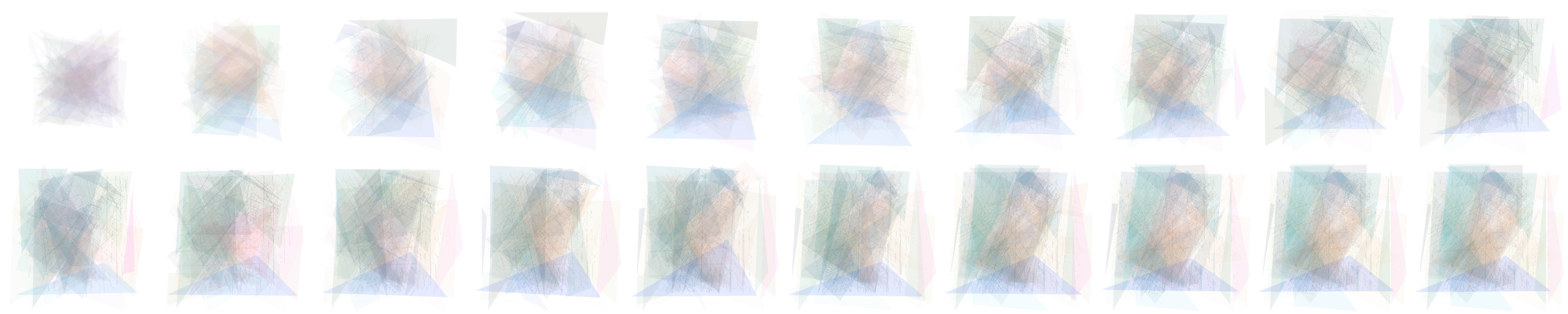} \\
        \midrule
        \includegraphics[width=\imgwidth]{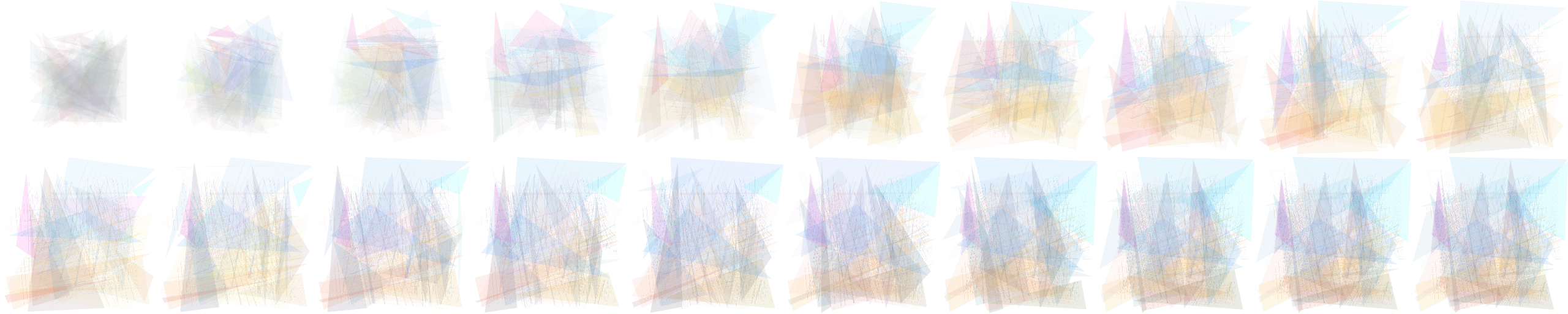} \\
    \bottomrule
    \end{tabular}
\end{subfigure}
\vskip -0.1cm
\caption{Evolution Strategies (non-gradient method) v.s. Differentiable renderer (gradient-based method) with fitting text with CLIP. Both settings are fitting 200 triangles to the target prompts. The details of evolution process is shown in the bottom half, where the upper group is for evolution strategy and the lower group is for differtiable renderer.}
\label{fig:es-clip-vs-diff-clip}
\vskip -0.9cm
\end{figure}

\textbf{Multiple Runs}.
Since the target is an abstract concept rather than a concrete image, our method is given much freedom in arranging the configuration of triangles, which means random initialization and noise in the optimization can lead to drastically different solutions. 
In Figure~\ref{fig:es-clip-multiple-runs}, we show 4 separate runs of our method on several text prompts, each using $50$ triangles with $2,000$ iterations, which is the same as previous examples.
As shown, our method creates distinctive abstractions aligned with human interpretation of language while being capable of producing diverse results from the same text prompt.
This, again, is a desired property for computer-assisted art creation, where human creators can be put ``in the loop'', not only poking around the text prompt but also picking the from multiple candidates produced by our method.

\textbf{Comparison with Gradient-based Optimization}.
With CLIP in mind, we are also interested in how our ES-based approach compares to the gradient-based optimization, especially since many existing works~\cite{wang2021bigsleep,samburtonking2021introduction,wang2021deepdaze,frans2021clipdraw} have proposed to leverage CLIP to guide the generations using gradients.
Arguably, this is a more challenging task due to the dynamic presented by two drastically different gradient dynamics by renderer and CLIP.
Usually, to make such kind of combination work ideally, more studies are required, which warrant a manuscript itself like~\cite{samburtonking2021introduction,frans2021clipdraw}.
Nonetheless, we have made a reasonably working version for comparison. 
Like fitting a target image, we implement the rendering process of composing triangles using nvdiffrast~\cite{laine2020modular}. 
In the forward pass, we render the canvas from parameters, feed the canvas to CLIP image encoder, and use Cosine distance between encoded image and encoded text prompt as a loss. Then we back-propagate all the way til the parameters of triangles to allow gradient-based optimization.
We use the same processing as mentioned in Section~\ref{sec:method}.

As shown in Figure~\ref{fig:es-clip-vs-diff-clip}, while both our ES method and the differentiable method produce images that are aligned with human interpretation of the text prompt,
ours produces more clear abstraction and clear boundaries between shapes and objects.
More interestingly, since ours represents an art style closely resembling abstract expressionism art, the difference between ours and the differentiable rendered is similar to that between post-impressionism and impressionism, where bolder geometric forms and colors are used.
Like the counterpart comparison in fitting a concrete image, we argue that such results are intrinsically rooted in the optimization mechanism, and our proposed method leads to a unique art style through our design choices.

\section{Discussion and Conclusion}

In this work, we revisit evolutionary algorithms for computational creativity by proposing to combine modern evolution strategies (ES) algorithms with the drawing primitives of triangles inspired by the minimalism art style. 
Our proposed method offers considerable improvements in both quality and efficiency compared to traditional genetic algorithms and is comparable to gradient-based methods.
Furthermore, we demonstrate that the ES algorithm could produce diverse, distinct geometric abstractions aligned with human interpretation of language and images.
Our finds suggests that ES method produce very different and sometimes better results compared to gradient based methods, arguably due to the intrinsical behavior of the optimization mechanism. 
However it remains an open problem to understand how in general setting ES method compares with gradient methods.
We expect future works investigate further into broader spectrum of art forms beyond the minimalism explored here.

Our dealing with evolutionary algorithms provides an insight into a different paradigm that can be applied to computational creativity.
Widely adopted gradient-based methods are fine-tuned for specific domains, i.e., diff rendering for edges, parameterized shapes, or data-driven techniques for rendering better textures. Each of the applications requires tunes and tweaks that are domain-specific and are hard to transfer.
In contrast, ES is agnostic to the domain, i.e., how the renderer works. 
We envision that ES-inspired approaches could potentially unify various domains with significantly less effort for adaption in the future.

\section{Acknowledgement}

We thank Toru Lin, Jerry Li, Yujin Tang, Yanghua Jin, Jesse Engel, Yifu Zhao for their comments and suggestions.
We specially thank Yanghua Jin for his kind help with nvdiffrast. The experiments in this work performed on multi-GPU Linux virtual machines provided by Google Cloud Platform.

\bibliographystyle{plainnat}
\bibliography{references}

\begin{thebibliography}{57}
\providecommand{\natexlab}[1]{#1}
\providecommand{\url}[1]{\texttt{#1}}
\expandafter\ifx\csname urlstyle\endcsname\relax
  \providecommand{\doi}[1]{doi: #1}\else
  \providecommand{\doi}{doi: \begingroup \urlstyle{rm}\Url}\fi

\bibitem[Alteredqualia(2008)]{alteredqualia2008evolutiongenetic}
Alteredqualia.
\newblock Evolution of mona lisa in javascript and canvas, 2008.
\newblock URL \url{https://alteredqualia.com/visualization/evolve/}.

\bibitem[Berg et~al.(2019)Berg, Berggren, Borgeteien, Jahren, Sajid, and
  Nichele]{berg2019evolved}
Joachim Berg, Nils Gustav~Andreas Berggren, Sivert~Allergodt Borgeteien,
  Christian Ruben~Alexander Jahren, Arqam Sajid, and Stefano Nichele.
\newblock Evolved art with transparent, overlapping, and geometric shapes.
\newblock In \emph{Symposium of the Norwegian AI Society}, pages 3--15.
  Springer, 2019.
\newblock URL \url{https://arxiv.org/abs/1904.06110}.

\bibitem[Bertoni(2002)]{bertoni2002minimalist}
Franco Bertoni.
\newblock \emph{Minimalist architecture}.
\newblock Birkhauser, 2002.
\newblock URL
  \url{https://www.worldcat.org/title/minimalist-architecture/oclc/469368801?loc=}.

\bibitem[Beyer(2001)]{beyer2001theory}
Hans-Georg Beyer.
\newblock \emph{The theory of evolution strategies}.
\newblock Springer Science \& Business Media, 2001.
\newblock URL \url{https://www.springer.com/gp/book/9783540672975}.

\bibitem[Beyer and Schwefel(2002)]{beyer2002evolution}
Hans-Georg Beyer and Hans-Paul Schwefel.
\newblock Evolution strategies--a comprehensive introduction.
\newblock \emph{Natural computing}, 1\penalty0 (1):\penalty0 3--52, 2002.

\bibitem[Cason(2016)]{cason2016}
Kenny Cason.
\newblock Genetic draw, 2016.
\newblock URL \url{https://github.com/kennycason/genetic_draw}.

\bibitem[Commons(2020)]{wiki:landscape}
Wikimedia Commons.
\newblock File:040 okertalsperre.jpg --- wikimedia commons the free media
  repository, 2020.
\newblock URL
  \url{https://commons.wikimedia.org/w/index.php?title=File:040_Okertalsperre.jpg&oldid=496749636}.
\newblock [Online; accessed 25-August-2021].

\bibitem[Dabrowski(2004)]{dabrowski2004geometric}
Magdalena Dabrowski.
\newblock Geometric abstraction, 2004.
\newblock URL \url{https://www.metmuseum.org/toah/hd/geab/hd_geab.htm}.

\bibitem[Fernando(2021)]{fernando2021royal}
Chrisantha Fernando.
\newblock Royal academy summer exhibition 2021 submission, 2021.
\newblock URL
  \url{https://www.chrisantha.co.uk/post/royal-academy-summer-exhibition-2021-submission}.

\bibitem[Fernando et~al.(2021)Fernando, Eslami, Alayrac, Mirowski, Banarse, and
  Osindero]{fernando2021generative}
Chrisantha Fernando, SM~Eslami, Jean-Baptiste Alayrac, Piotr Mirowski, Dylan
  Banarse, and Simon Osindero.
\newblock Generative art using neural visual grammars and dual encoders.
\newblock \emph{arXiv preprint arXiv:2105.00162}, 2021.
\newblock URL \url{https://arxiv.org/abs/2105.00162}.

\bibitem[Fogleman(2016)]{fogleman2016}
Michael Fogleman.
\newblock Primitive pictures, 2016.
\newblock URL \url{https://github.com/fogleman/primitive}.

\bibitem[Frans et~al.(2021)Frans, Soros, and Witkowski]{frans2021clipdraw}
Kevin Frans, LB~Soros, and Olaf Witkowski.
\newblock Clipdraw: Exploring text-to-drawing synthesis through language-image
  encoders.
\newblock \emph{arXiv preprint arXiv:2106.14843}, 2021.
\newblock URL \url{https://arxiv.org/abs/2106.14843}.

\bibitem[Galatolo et~al.(2021)Galatolo, Cimino, and
  Vaglini]{galatolo2021generating}
Federico~A Galatolo, Mario~GCA Cimino, and Gigliola Vaglini.
\newblock Generating images from caption and vice versa via clip-guided
  generative latent space search.
\newblock \emph{arXiv preprint arXiv:2102.01645}, 2021.
\newblock URL \url{https://arxiv.org/abs/2102.01645}.

\bibitem[Ganin et~al.(2018)Ganin, Kulkarni, Babuschkin, Eslami, and
  Vinyals]{ganin2018synthesizing}
Yaroslav Ganin, Tejas Kulkarni, Igor Babuschkin, SM~Ali Eslami, and Oriol
  Vinyals.
\newblock Synthesizing programs for images using reinforced adversarial
  learning.
\newblock In \emph{International Conference on Machine Learning}, pages
  1666--1675. PMLR, 2018.
\newblock URL \url{http://proceedings.mlr.press/v80/ganin18a.html}.

\bibitem[Gonsalves(2021)]{gonsalves2021impressionist}
Robert~A. Gonsalves.
\newblock Ganscapes: Using ai to create new impressionist paintings, 2021.
\newblock URL
  \url{https://towardsdatascience.com/ganscapes-using-ai-to-create-new-impressionist-paintings-d6af1cf94c56}.

\bibitem[Gregor et~al.(2015)Gregor, Danihelka, Graves, Rezende, and
  Wierstra]{gregor2015draw}
Karol Gregor, Ivo Danihelka, Alex Graves, Danilo Rezende, and Daan Wierstra.
\newblock Draw: A recurrent neural network for image generation.
\newblock In \emph{International Conference on Machine Learning}, pages
  1462--1471. PMLR, 2015.
\newblock URL \url{http://proceedings.mlr.press/v37/gregor15.html}.

\bibitem[Ha(2017)]{ha2017evolving}
David Ha.
\newblock Evolving stable strategies.
\newblock \emph{blog.otoro.net}, 2017.
\newblock URL
  \url{http://blog.otoro.net/2017/11/12/evolving-stable-strategies/}.

\bibitem[Ha and Eck(2017)]{ha2017neural}
David Ha and Douglas Eck.
\newblock A neural representation of sketch drawings.
\newblock \emph{arXiv preprint arXiv:1704.03477}, 2017.
\newblock URL \url{https://arxiv.org/abs/1704.03477}.

\bibitem[Harvey(2009)]{harvey2009microbial}
Inman Harvey.
\newblock The microbial genetic algorithm.
\newblock In \emph{European conference on artificial life}, pages 126--133.
  Springer, 2009.
\newblock URL \url{http://users.sussex.ac.uk/~inmanh/MicrobialGA_ECAL2009.pdf}.

\bibitem[Hochreiter and Schmidhuber(1997)]{hochreiter1997long}
Sepp Hochreiter and J{\"u}rgen Schmidhuber.
\newblock Long short-term memory.
\newblock \emph{Neural computation}, 9\penalty0 (8):\penalty0 1735--1780, 1997.
\newblock URL
  \url{https://direct.mit.edu/neco/article/9/8/1735/6109/Long-Short-Term-Memory}.

\bibitem[Huang and Canny(2019)]{huang2019sketchforme}
Forrest Huang and John~F Canny.
\newblock Sketchforme: Composing sketched scenes from text descriptions for
  interactive applications.
\newblock In \emph{Proceedings of the 32nd Annual ACM Symposium on User
  Interface Software and Technology}, pages 209--220, 2019.
\newblock URL \url{https://dl.acm.org/doi/10.1145/3332165.3347878}.

\bibitem[Huang et~al.(2020)Huang, Schoop, Ha, and Canny]{huang2020scones}
Forrest Huang, Eldon Schoop, David Ha, and John Canny.
\newblock Scones: towards conversational authoring of sketches.
\newblock In \emph{Proceedings of the 25th International Conference on
  Intelligent User Interfaces}, pages 313--323, 2020.
\newblock URL \url{https://dl.acm.org/doi/10.1145/3377325.3377485}.

\bibitem[Huang et~al.(2019)Huang, Heng, and Zhou]{huang2019learning}
Zhewei Huang, Wen Heng, and Shuchang Zhou.
\newblock Learning to paint with model-based deep reinforcement learning.
\newblock In \emph{Proceedings of the IEEE/CVF International Conference on
  Computer Vision}, pages 8709--8718, 2019.
\newblock URL
  \url{https://openaccess.thecvf.com/content_ICCV_2019/html/Huang_Learning_to_Paint_With_Model-Based_Deep_Reinforcement_Learning_ICCV_2019_paper.html}.

\bibitem[Jia et~al.(2021)Jia, Yang, Xia, Chen, Parekh, Pham, Le, Sung, Li, and
  Duerig]{jia2021scaling}
Chao Jia, Yinfei Yang, Ye~Xia, Yi-Ting Chen, Zarana Parekh, Hieu Pham, Quoc~V
  Le, Yunhsuan Sung, Zhen Li, and Tom Duerig.
\newblock Scaling up visual and vision-language representation learning with
  noisy text supervision.
\newblock \emph{arXiv preprint arXiv:2102.05918}, 2021.
\newblock URL \url{https://arxiv.org/abs/2102.05918}.

\bibitem[Johansson(2008)]{johansson2008genetic}
Roger Johansson.
\newblock Genetic programming: Evolution of mona lisa, 2008.
\newblock URL
  \url{https://rogerjohansson.blog/2008/12/07/genetic-programming-evolution-of-mona-lisa/}.

\bibitem[Kato et~al.(2020)Kato, Beker, Morariu, Ando, Matsuoka, Kehl, and
  Gaidon]{kato2020differentiable}
Hiroharu Kato, Deniz Beker, Mihai Morariu, Takahiro Ando, Toru Matsuoka, Wadim
  Kehl, and Adrien Gaidon.
\newblock Differentiable rendering: A survey.
\newblock \emph{arXiv preprint arXiv:2006.12057}, 2020.
\newblock URL \url{https://arxiv.org/abs/2006.12057}.

\bibitem[Kingma and Ba(2014)]{kingma2014adam}
Diederik~P Kingma and Jimmy Ba.
\newblock Adam: A method for stochastic optimization.
\newblock \emph{arXiv preprint arXiv:1412.6980}, 2014.
\newblock URL \url{https://arxiv.org/abs/1412.6980}.

\bibitem[Kolmogorov(1965)]{kolmogorov1965three}
Andrei Kolmogorov.
\newblock Three approaches to the quantitative definition of information.
\newblock \emph{Problems of information transmission}, 1\penalty0 (1):\penalty0
  1--7, 1965.
\newblock URL
  \url{https://www.tandfonline.com/doi/abs/10.1080/00207166808803030?journalCode=gcom20}.

\bibitem[Kuiper(2021)]{kuiper2021modernism}
Kathleen Kuiper.
\newblock Modernism, 2021.
\newblock URL \url{https://www.britannica.com/art/Modernism-art}.

\bibitem[Laine et~al.(2020)Laine, Hellsten, Karras, Seol, Lehtinen, and
  Aila]{laine2020modular}
Samuli Laine, Janne Hellsten, Tero Karras, Yeongho Seol, Jaakko Lehtinen, and
  Timo Aila.
\newblock Modular primitives for high-performance differentiable rendering.
\newblock \emph{ACM Transactions on Graphics (TOG)}, 39\penalty0 (6):\penalty0
  1--14, 2020.
\newblock URL \url{https://arxiv.org/abs/2011.03277}.

\bibitem[Li et~al.(2020)Li, Luk{\'a}{\v{c}}, Gharbi, and
  Ragan-Kelley]{li2020differentiable}
Tzu-Mao Li, Michal Luk{\'a}{\v{c}}, Micha{\"e}l Gharbi, and Jonathan
  Ragan-Kelley.
\newblock Differentiable vector graphics rasterization for editing and
  learning.
\newblock \emph{ACM Transactions on Graphics (TOG)}, 39\penalty0 (6):\penalty0
  1--15, 2020.
\newblock URL \url{https://people.csail.mit.edu/tzumao/diffvg/}.

\bibitem[Lindenmayer(1968)]{lindenmayer1968mathematical}
Aristid Lindenmayer.
\newblock Mathematical models for cellular interactions in development i.
  filaments with one-sided inputs.
\newblock \emph{Journal of theoretical biology}, 18\penalty0 (3):\penalty0
  280--299, 1968.
\newblock URL
  \url{https://www.sciencedirect.com/science/article/abs/pii/0022519368900799}.

\bibitem[Liu et~al.(2021)Liu, Lin, He, Li, Deng, Li, Ding, and
  Wang]{liu2021paint}
Songhua Liu, Tianwei Lin, Dongliang He, Fu~Li, Ruifeng Deng, Xin Li, Errui
  Ding, and Hao Wang.
\newblock Paint transformer: Feed forward neural painting with stroke
  prediction.
\newblock \emph{arXiv preprint arXiv:2108.03798}, 2021.
\newblock URL \url{https://arxiv.org/abs/2108.03798}.

\bibitem[Lopes et~al.(2019)Lopes, Ha, Eck, and Shlens]{lopes2019learned}
Raphael~Gontijo Lopes, David Ha, Douglas Eck, and Jonathon Shlens.
\newblock A learned representation for scalable vector graphics.
\newblock In \emph{Proceedings of the IEEE/CVF International Conference on
  Computer Vision}, pages 7930--7939, 2019.
\newblock URL
  \url{https://openaccess.thecvf.com/content_ICCV_2019/html/Lopes_A_Learned_Representation_for_Scalable_Vector_Graphics_ICCV_2019_paper.html}.

\bibitem[Malkevitchn(2003)]{malkevitch2003mathematics}
Joseph Malkevitchn.
\newblock Mathematics and art, 2003.
\newblock URL
  \url{https://www.ams.org/publicoutreach/feature-column/fcarc-art1}.

\bibitem[Mellor et~al.(2019)Mellor, Park, Ganin, Babuschkin, Kulkarni,
  Rosenbaum, Ballard, Weber, Vinyals, and Eslami]{mellor2019unsupervised}
John~FJ Mellor, Eunbyung Park, Yaroslav Ganin, Igor Babuschkin, Tejas Kulkarni,
  Dan Rosenbaum, Andy Ballard, Theophane Weber, Oriol Vinyals, and SM~Eslami.
\newblock Unsupervised doodling and painting with improved spiral.
\newblock \emph{arXiv preprint arXiv:1910.01007}, 2019.
\newblock URL \url{https://arxiv.org/abs/1910.01007}.

\bibitem[Modern(2018)]{tate_minimalism}
Tate Modern.
\newblock Minimalism, 2018.
\newblock URL \url{https://www.tate.org.uk/art/art-terms/m/minimalism}.

\bibitem[Nakano(2019)]{nakano2019neural}
Reiichiro Nakano.
\newblock Neural painters: A learned differentiable constraint for generating
  brushstroke paintings.
\newblock \emph{arXiv preprint arXiv:1904.08410}, 2019.
\newblock URL \url{https://arxiv.org/abs/1904.08410}.

\bibitem[Paauw and Van~den Berg(2019)]{paauw2019paintings}
Misha Paauw and Daan Van~den Berg.
\newblock Paintings, polygons and plant propagation.
\newblock In \emph{International Conference on Computational Intelligence in
  Music, Sound, Art and Design (Part of EvoStar)}, pages 84--97. Springer,
  2019.
\newblock URL
  \url{https://link.springer.com/chapter/10.1007/978-3-030-16667-0_6}.

\bibitem[Paul(2004)]{paul2004abstract}
Stella Paul.
\newblock Abstract expressionism, 2004.
\newblock URL \url{https://www.metmuseum.org/toah/hd/abex/hd_abex.htm}.

\bibitem[Radford et~al.(2021)Radford, Kim, Hallacy, Ramesh, Goh, Agarwal,
  Sastry, Askell, Mishkin, Clark, et~al.]{radford2021learning}
Alec Radford, Jong~Wook Kim, Chris Hallacy, Aditya Ramesh, Gabriel Goh,
  Sandhini Agarwal, Girish Sastry, Amanda Askell, Pamela Mishkin, Jack Clark,
  et~al.
\newblock Learning transferable visual models from natural language
  supervision.
\newblock \emph{arXiv preprint arXiv:2103.00020}, 2021.
\newblock URL \url{https://arxiv.org/abs/2103.00020}.

\bibitem[Rewald(2014)]{rewald2014heilbrunn}
Sabine Rewald.
\newblock Heilbrunn timeline of art history: Cubism.
\newblock \emph{The Metropolitan Museum of Art}, 2014.
\newblock URL \url{https://www.metmuseum.org/toah/hd/cube/hd_cube.htm}.

\bibitem[Rose(1965)]{rose1965abc}
Barbara Rose.
\newblock Abc art.
\newblock \emph{Art in America}, 53\penalty0 (5):\penalty0 57--69, 1965.
\newblock URL
  \url{https://www.artnews.com/art-in-america/features/abc-art-barbara-rose-1234580665/}.

\bibitem[Salimans et~al.(2017)Salimans, Ho, Chen, Sidor, and
  Sutskever]{salimans2017evolution}
Tim Salimans, Jonathan Ho, Xi~Chen, Szymon Sidor, and Ilya Sutskever.
\newblock Evolution strategies as a scalable alternative to reinforcement
  learning.
\newblock \emph{arXiv preprint arXiv:1703.03864}, 2017.
\newblock URL \url{https://arxiv.org/abs/1703.03864}.

\bibitem[@samburtonking(2021)]{samburtonking2021introduction}
@samburtonking.
\newblock Introduction to vqgan+clip, 2021.
\newblock URL
  \url{https://docs.google.com/document/d/1Lu7XPRKlNhBQjcKr8k8qRzUzbBW7kzxb5Vu72GMRn2E/edit}.

\bibitem[Schmidhuber(1997)]{schmidhuber1997low}
J{\"u}rgen Schmidhuber.
\newblock Low-complexity art.
\newblock \emph{Leonardo}, 30\penalty0 (2):\penalty0 97--103, 1997.
\newblock URL \url{https://www.idsia.ch/~juergen/locoart/locoart.html}.

\bibitem[Sehnke et~al.(2010)Sehnke, Osendorfer, R{\"u}ckstie{\ss}, Graves,
  Peters, and Schmidhuber]{sehnke2010parameter}
Frank Sehnke, Christian Osendorfer, Thomas R{\"u}ckstie{\ss}, Alex Graves, Jan
  Peters, and J{\"u}rgen Schmidhuber.
\newblock Parameter-exploring policy gradients.
\newblock \emph{Neural Networks}, 23\penalty0 (4):\penalty0 551--559, 2010.
\newblock URL \url{https://people.idsia.ch//~juergen/nn2010.pdf}.

\bibitem[Shahrabi(2020)]{shahrabi2020}
Shahriar Shahrabi.
\newblock Procedural paintings with genetic evolution algorithm, 2020.
\newblock URL \url{https://github.com/IRCSS/Procedural-painting}.

\bibitem[Studios"(2019)]{sizigi2019waifu}
"Sizigi Studios".
\newblock Waifu labs, 2019.
\newblock URL \url{https://waifulabs.com/}.

\bibitem[Tate(2021)]{tate2021process}
Tate.
\newblock Process art, 2021.
\newblock URL \url{https://www.tate.org.uk/art/art-terms/p/process-art}.

\bibitem[Toklu(2020)]{toklu2008pgpelib}
Nihat~Engin Toklu.
\newblock Pgpelib.
\newblock \url{https://github.com/nnaisense/pgpelib}, 2020.
\newblock URL \url{https://github.com/nnaisense/pgpelib}.

\bibitem[Toklu et~al.(2020)Toklu, Liskowski, and Srivastava]{toklu2020clipup}
Nihat~Engin Toklu, Pawe{\l} Liskowski, and Rupesh~Kumar Srivastava.
\newblock Clipup: A simple and powerful optimizer for distribution-based policy
  evolution.
\newblock In \emph{International Conference on Parallel Problem Solving from
  Nature}, pages 515--527. Springer, 2020.
\newblock URL \url{https://arxiv.org/abs/2008.02387}.

\bibitem[Verostko(1994)]{verostko1994algorithmic}
Roman Verostko.
\newblock Algorithmic art, 1994.
\newblock URL \url{http://www.verostko.com/algorithm.html}.

\bibitem[Wang(2021{\natexlab{a}})]{wang2021bigsleep}
Phil Wang.
\newblock Big sleep: A simple command line tool for text to image generation,
  using openai's clip and a biggan, 2021{\natexlab{a}}.
\newblock URL \url{https://github.com/lucidrains/big-sleep}.

\bibitem[Wang(2021{\natexlab{b}})]{wang2021deepdaze}
Phil Wang.
\newblock Deep daze: A simple command line tool for text to image generation
  using openai's clip and siren (implicit neural representation network),
  2021{\natexlab{b}}.
\newblock URL \url{https://github.com/lucidrains/deep-daze}.

\bibitem[White(2019)]{white2019shared}
Tom White.
\newblock Shared visual abstractions.
\newblock \emph{arXiv preprint arXiv:1912.04217}, 2019.
\newblock URL \url{https://arxiv.org/abs/1912.04217}.

\bibitem[Zheng et~al.(2019)Zheng, Jiang, and Huang]{zheng2018strokenet}
Ningyuan Zheng, Yifan Jiang, and Dingjiang Huang.
\newblock Strokenet: A neural painting environment.
\newblock In \emph{International Conference on Learning Representations}, 2019.
\newblock URL \url{https://openreview.net/forum?id=HJxwDiActX}.

\end{thebibliography}

\end{document}